\pdfoutput=1
\documentclass[11pt]{article}

\usepackage{acl}
\usepackage{times}
\usepackage{latexsym}

\usepackage[T1]{fontenc}
\usepackage[utf8]{inputenc}

\usepackage{microtype}

\usepackage{inconsolata}

\usepackage{graphicx}
\usepackage{subcaption}
\usepackage{amsmath}
\usepackage{booktabs}
\usepackage{multirow}
\usepackage{hyperref}
\usepackage{mathtools}
\usepackage{hyperref}       % hyperlinks
\usepackage{url}            % simple URL typesetting
\usepackage{booktabs}       % professional-quality tables
\usepackage{amsfonts}       % blackboard math symbols
\usepackage{nicefrac}       % compact symbols for 1/2, etc.
\usepackage{microtype}      % microtypography
\usepackage{xcolor}         % colors
\usepackage{pifont}
\usepackage{float} 
\usepackage{booktabs}
\usepackage{caption}
\usepackage[ruled,vlined,linesnumbered]{algorithm2e}

\newcommand{\cmark}{\textcolor{green!70!black}{\ding{51}}} % ✔
\newcommand{\xmark}{\textcolor{red}{\ding{55}}}
\title{AdaTP: Attention-Debiased Token Pruning for \\ Video Large Language Models}

\author{
  Fengyuan Sun\textsuperscript{1}\thanks{~~Equal contribution.} \quad
  Leqi Shen\textsuperscript{1}\footnotemark[1] \quad
  Hui Chen\textsuperscript{2} \\
  \textbf{Sicheng Zhao\textsuperscript{2}} \quad
  \textbf{Jungong Han\textsuperscript{3}} \quad
  \textbf{Guiguang Ding\textsuperscript{1,2}} \\
  \textsuperscript{1}School of Software, Tsinghua University \quad
  \textsuperscript{2}BNRist, Tsinghua University \\
  \textsuperscript{3}Department of Automation, Tsinghua University \\
}

\begin{document}
\maketitle
\begin{abstract}
Video Large Language Models (Video LLMs) have achieved remarkable results in video understanding tasks. However, they often suffer from heavy computational overhead due to the large number of visual tokens generated from multiple video frames. Existing visual token compression methods often rely on attention scores from language models as guidance. However, these scores exhibit inherent biases: global bias reflects a tendency to focus on the two ends of the visual token sequence, while local bias leads to an over-concentration on the same spatial positions across different frames. To address the issue of attention bias, we propose \textbf{A}ttention-\textbf{D}ebi\textbf{a}sed \textbf{T}oken \textbf{P}runing for Video Large Language Models (\textbf{AdaTP}), a novel token pruning pipeline for Video LLMs. AdaTP integrates two dedicated debiasing modules into the pipeline, targeting global attention bias and local attention bias, respectively. Without the need for additional training, our method significantly reduces the computational overhead of Video LLMs while retaining the performance of vanilla models. Extensive evaluation shows that AdaTP achieves state-of-the-art performance in various commonly used video understanding benchmarks. In particular, on LLaVA-OneVision-7B, AdaTP maintains performance without degradation while using only up to 27.3\% FLOPs compared to the vanilla model. Our code will be released soon.

% deleted
% Specifically, the global debiasing module adaptively partitions each video into segments containing varying numbers of frames, identifies critical segments guided by textual information; while local debiasing module performs adaptive token pruning on each segment independently leveraging attention scores.
\end{abstract}

\section{Introduction}

Video large language models (Video LLMs)~\cite{Li2024LLaVAOneVision, lin2023video, wang2024qwen2, zhang2024video, maaz2023video} have shown potential and effectiveness in video comprehension domain. Building upon architectures of previous Multi-Modal Large Language Models~\cite{li2023blip, liu2024llavanext, zhu2023minigpt, team2023gemini, alayrac2022flamingo, liu2024improved, liu2023visual} designed primarily for image tasks, Video LLMs transforms raw videos into sequences of visual tokens in frame-by-frame manner. After alignment, these visual tokens are fed into the language model alongside text tokens.

\begin{figure*}[htbp]
  \centering

  \begin{subfigure}[b]{\linewidth}
    \centering
    \includegraphics[width=\linewidth]{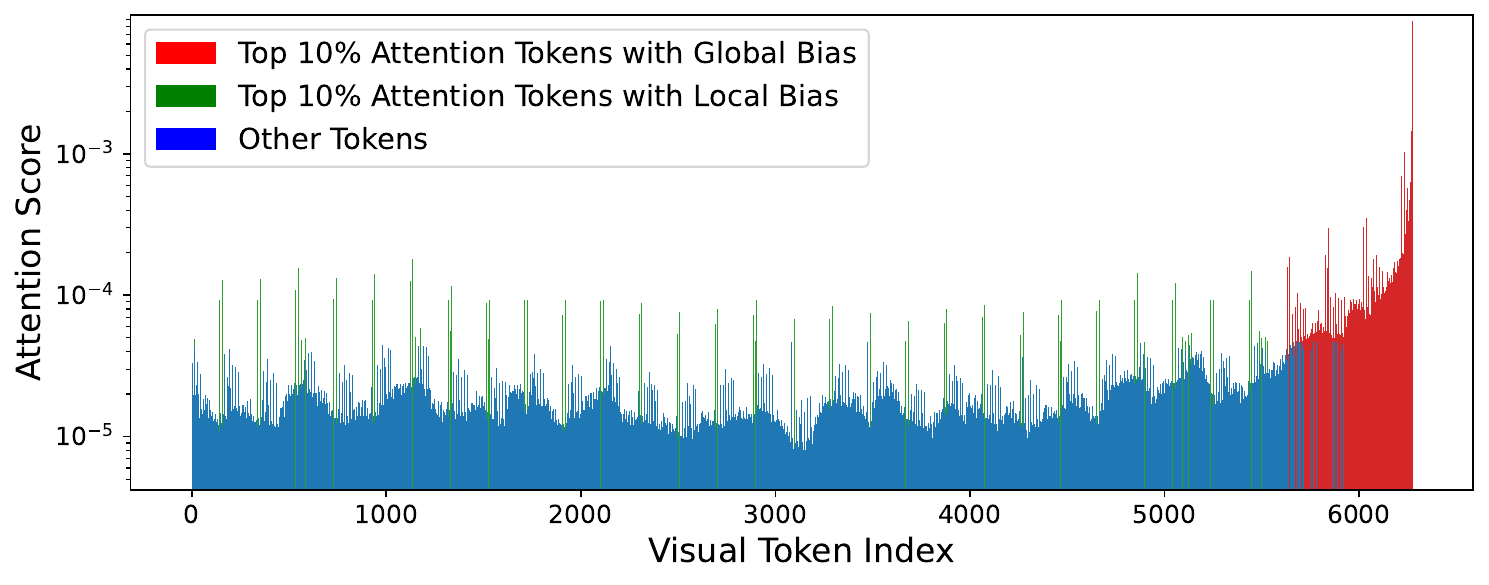}
    % \caption{Global attention distribution in Layer 1.}
    % \label{fig:heatmap1}
  \end{subfigure}

  \caption{Visualization of attention bias. The visual tokens with top-10\% attention scores are highlighted in red and green, where red tokens represent global attention bias, and green tokens represent local attention bias.}
  \label{fig:combined_heatmaps_frames}
\end{figure*}

Video LLMs face significant computational challenge, which can be mainly attributed to the large number of visual tokens and the quadratic complexity of attention mechanism. To mitigate this, some previous works~\cite{chen2024image, shang2024llava, wang2024cls, yang2024visionzip, shen2024tempme, tao2024dycoke, shen2025fastvid, shen2025llava} have sought to alleviate this through visual token compression, typically leveraging attention scores to guide token pruning or merging. However, these works often fail to accurately utilize attention scores for token compression process. As a result, the selected tokens may fail to retain a comprehensive representation of the visual content, and thus lead to obvious performance degradation.

In this work, we first identify that the attention scores in the language models of Video LLMs are inherently biased, which can be characterized from two perspectives:

\textbf{Global attention bias}. Visual tokens with high attention scores tend to cluster at the beginning or end of the sequence, potentially overlooking intermediate important visual information.

\textbf{Local attention bias}. We conduct frame-wise analysis, and discovers that in shallow layers, attention scores are disproportionately concentrated on a few spatial positions, leading to over-retention of those tokens and reduced visual diversity.

To provide a more intuitive understanding of attention biases, we visualize the average attention distribution in the LLaVA-OneVision-7B model~\cite{Li2024LLaVAOneVision}. The visualization is computed across all samples in the VideoMME~\cite{fu2024video} dataset and shown in Fig.~\ref{fig:combined_heatmaps_frames}. Visualization results from other layers can be found in the Appendix~\ref{subsec:more-vis-layers}. In Fig.~\ref{fig:combined_heatmaps_frames}, we can intuitively see that the red bars, which represent high attention scores, are concentrated in the latter part of the sequence, showing global attention bias; the green bars, exhibit a periodic pattern throughout the entire sequence, representing local attention bias.

To effectively leverage attention scores to guide visual token compression, we argue that it is essential to address the aforementioned biases. In this paper, we propose targeted modules to mitigate global and local attention bias respectively. Building upon these insights, we introduce Attention-Debiased Token Pruning for Video Large Language Models (\textbf{AdaTP}), a training-free and plug-and-play acceleration pipeline for Video LLMs. To be specific, we first partition adjacent similar frames into coherent segments, then introduce two debiasing modules: the Global Debiasing Module helps the model focus on semantically meaningful segments by identifying visual contents that are closely related to textual information; the Local Debiasing Module performs token pruning adaptively and independently within each segment, which encourages the preservation of diverse visual content and effectively reduces intra-segment redundancy.

We integrate our AdaTP pipeline into two types of representative Video LLM architectures: LLaVA-OneVision~\cite{Li2024LLaVAOneVision} and LLaVA-Video~\cite{lin2023video}. Experimental results show that our method achieves state-of-the-art results on several video understanding benchmarks including VideoMME~\cite{fu2024video}, LongVideoBench~\cite{wu2024longvideobench}, and MLVU~\cite{zhou2024mlvu}, which are widely adopted to assess the comprehension ability of Video LLMs, and contain diverse video types. On LLaVA-OneVision-7B, our method achieves comparable performance to the original model while keeping only 27.3\% of its FLOPs.

Our main contributions are as follows:

% 1. We reveal and thoroughly analyze the attention bias phenomenon in Video LLMs from both global and local perspectives.
    
% 2. We propose AdaTP, a novel token pruning framework for Video LLMs. AdaTP mitigates attention biases and leverages debiased attention scores as an effective guidance for visual token compression, thus reduced computational burden without sacrificing models' performance.
    
% 3. Experimental results demonstrate that AdaTP achieves state-of-the-art performance on diverse video comprehension benchmarks without requiring additional training.

\begin{enumerate}
    \item We reveal and thoroughly analyze the attention bias phenomenon in Video LLMs from both global and local perspectives.
    
    \item We propose AdaTP, a novel token pruning framework for Video LLMs. AdaTP mitigates attention biases and leverages debiased attention scores as an effective guidance for visual token compression, thus reduced computational burden without sacrificing models' performance.
    
    \item Experimental results demonstrate that AdaTP achieves state-of-the-art performance on diverse video comprehension benchmarks without requiring additional training.
\end{enumerate}

\section{Related Works}

\subsection{Video Large Language Models}
With the success of Multi-modal Large Language Models (MLLMs), increasing efforts have been made to extend their capabilities to video understanding. Video-LLaVA~\cite{lin2023video} aligns images and videos, allowing language models to learn a unified visual representation, enabling language models to understand images and videos content simultaneously. Qwen-2-VL~\cite{wang2024qwen2} extends dynamic resolution and updates mRoPE in the time dimension, enhancing the model's ability to learn from temporal sequence. PLLaVA~\cite{xu2024pllava} applies pooling strategy to smooth the feature distribution along the temporal dimension and extended Image LLMs. LLaVA-OneVision~\cite{Li2024LLaVAOneVision} unifies image and video under a single framework for consistency and effectiveness. LLaVA-Video~\cite{zhang2024video} extends visual instruction tuning to videos, training model with synthetic dataset for video instruction-following, and introduces newline tokens for each frame to distinguish spatial and temporal positions. InternVL-2.5~\cite{chen2024expanding} uses pixel unshuffle operation to reduce the number of visual tokens.

LLaVA-OneVision~\cite{Li2024LLaVAOneVision} and LLaVA-Video~\cite{zhang2024video} are two of the most widely used and representative Video LLMs, thus we choose to evaluate our method on these two model architectures.

\subsection{Visual token compression}
Token compression methods have been proposed to accelerate visual encoder and MLLM's inference speed. In earlier years, several works~\cite{rao2021dynamicvit, yin2022vit, liang2022not, marin2021token} were done to explore token compression for Vision Transformers (ViT). Recent studies propose visual token compression methods to accelerate inference in MLLMs and Video LLMs. FastV~\cite{chen2024image} reveals the high redundancy of visual tokens in language models and leverages attention scores from it to guide the pruning process. LLaVA-PruMerge~\cite{shang2024llava} and several subsequent works~\cite{wang2024cls, zhang2024cls} perform token pruning and merging based on the attention scores of [CLS] token from the visual encoder. VisionZip~\cite{yang2024visionzip} selects a small number of key visual tokens based on the attention information from the visual encoder, and applies a merging strategy to retain the remaining information. Dycoke~\cite{tao2024dycoke} is specifically tailored for Video LLMs, which performs fixed-length segmentation of videos before feeding them into the LLM, removes and merges redundant parts, and further prunes tokens during the decoding stage of the LLM.

Most of the aforementioned methods rely on attention scores to guide visual token compression and have achieved promising results. However, their overall effectiveness remains limited. These methods often suffer from notable performance degradation due to attention bias. In contrast, our method mitigates attention bias through a targeted debiasing mechanism, enabling more efficient and effective visual token compression.

\section{Methodology}

% In this section, we present our AdaTP pipeline for Video LLMs inference acceleration. We first introduce the preliminaries of Video LLM's inference process in Section \ref{subsec:preliminaries}. We then discuss our observations on attention bias phenomenon in Section \ref{subsec:observation}, constitutes the main issue we aim to address. Based on these findings, in Section \ref{subsec:pipeline}, we propose our new pipeline named AdaTP for Video LLM's inference acceleration by incorporating attention bias correction.

% First-level headings should be in 12-point type.

\subsection{Preliminaries}
\label{subsec:preliminaries}
\textbf{Video LLM's Inference.} Video LLM's inference process can be divided into the following steps: a video is processed by visual encoder (such as CLIP~\cite{Radford2021LearningTV} and SigLIP~\cite{zhai2023sigmoid}) in frame-by-frame paradigm. After alignment through a projector, the visual tokens are concatenated with tokenized text tokens and system prompt tokens, serve together as the input to the LLM. We denote $VE(\cdot)$ as the visual encoder that processes the video $v$, and $P(\cdot)$ as the projector alignment process. The processed visual tokens $P(VE(v))$ can be represented as:
\begin{equation}
X_v^0 = P(VE(v)) \in \mathbb{R}^{nc \times d},
\label{eq0}
\end{equation}
where $X_v^0$ is the output of the visual projector and serves as the input to the 0-th layer of the LLM, while $X_v^l$ denotes the input to the $l$-th layer. $X_{v}^{l}$ can also be represented as $[X_{v_{f_1}}^l, \ldots, X_{v_{f_n}}^l]$ in frame-by-frame manner, and $X_{v_{f_i}}^l(i = 1, 2, \ldots, n)$ is denoted as visual tokens of the $i$-th frame from the video in the $l$-th layer, where $n$ is the frame count. $c$ is the number of tokens within any single frame, and $d$ is the hidden dimension of visual tokens.

\textbf{Attention score in LLM as token importance metric.} In Video LLMs, attention scores produced by LLMs can serve as a metric for token importance. In the self-attention module of the $l$-th layer in language models, by averaging across all attention heads, we get attention map $A^l \in \mathbb{R}^{L \times L}$. We extract the attention scores from text tokens to visual tokens by averaging the rows of the attention matrix $A^l$ corresponding to text tokens and slicing the columns corresponding to visual tokens, resulting in $s^l \in \mathbb{R}^{nc}$.
In the $(l+1)$-th layer, these scores can be used to guide the token compression process. However, directly selecting the visual tokens with the highest attention scores leads to significant performance degradation, as will be discussed in the next subsection.

\subsection{Observation: Attention bias phenomenon}
\label{subsec:observation}

\begin{figure}[htbp]
  \centering

  \begin{subfigure}[b]{\linewidth}
    \centering
    \includegraphics[width=\linewidth]{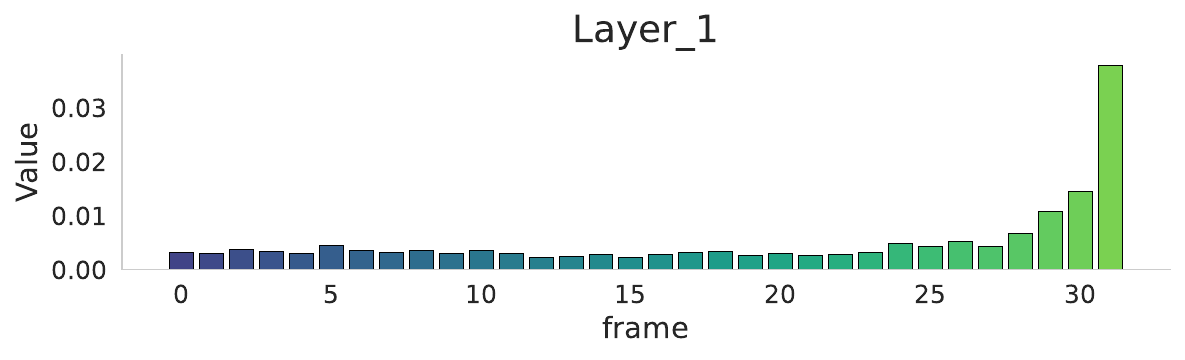}
    % \caption{Global attention distribution in Layer 1.}
    \label{fig:heatmap1}
  \end{subfigure}

  \vspace{-0.5cm} % 可根据需要调节两图之间的间距

  \begin{subfigure}[b]{\linewidth}
    \centering
    \includegraphics[width=\linewidth]{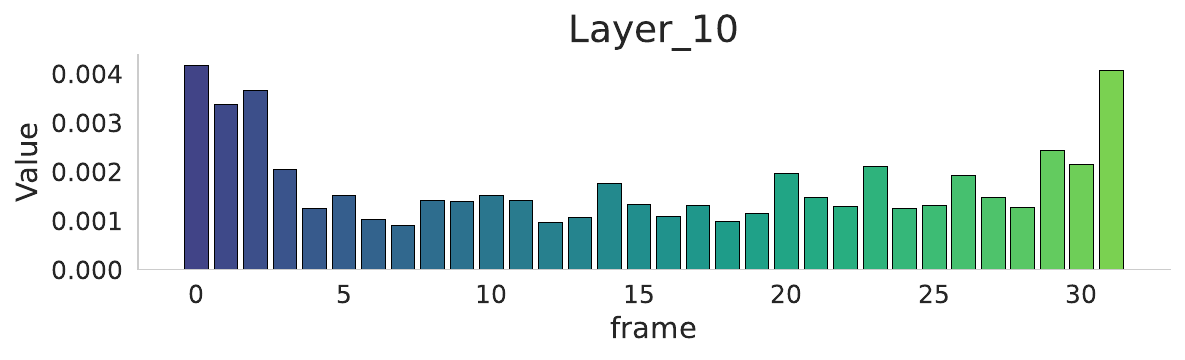}
    % \caption{Global attention distribution in Layer 10.}
    \label{fig:heatmap2}
  \end{subfigure}

  \caption{Visualization of global attention bias in Layer 1 and 10. Each bar represents the sum of attention scores of visual tokens within a single frame.}
  \label{fig:combined_heatmaps}
\end{figure}

Many existing works~\cite{zhang2023h2o, xiao2023efficient, chen2024image, tao2024dycoke} leverage attention scores to guide token pruning. However, Such strategies are actually suboptimal due to the presence of attention bias, leaving substantial room for improvement. The two parts of attention biases are stated as follows:

\textbf{\textit{Observation 1.}} \textbf{Global attention bias.} Ideally, to serve as a visual token importance metric, for specific text-video pairs, attention scores should be concentrated on tokens from frames that are highly relevant to the text. However, in practice, in almost all layers, these tokens are predominantly located in the beginning and the final portion of the sequence, regardless of the content of text-video pair. In Fig.~\ref{fig:combined_heatmaps}, each bar represents the sum of attention scores of all visual tokens within a single video frame. This figure clearly illustrates the presence of global bias, as the attention scores are concentrated at the first and last few frames, while the middle frames receive significantly less attention. We take attention distribution of layer 1 as an example, among the tokens with the top 10\% attention scores, 86.8\% are concentrated in the last 4 frames out of total 32 frames, which clearly shows the global attention bias.

\begin{figure}[htbp]
  \centering

  \begin{subfigure}[b]{0.48\linewidth}
    \centering
    \includegraphics[width=\linewidth]{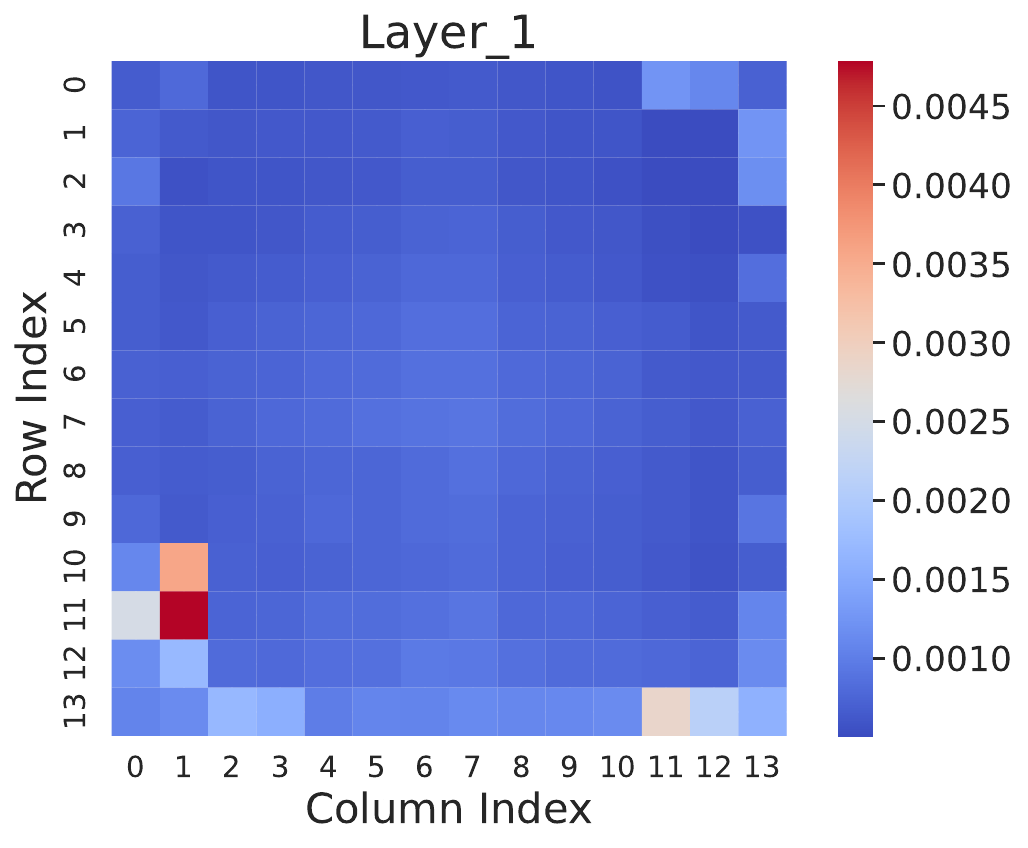}
  \end{subfigure}
  \hspace{0.00\linewidth}
  \begin{subfigure}[b]{0.48\linewidth}
    \centering
    \includegraphics[width=\linewidth]{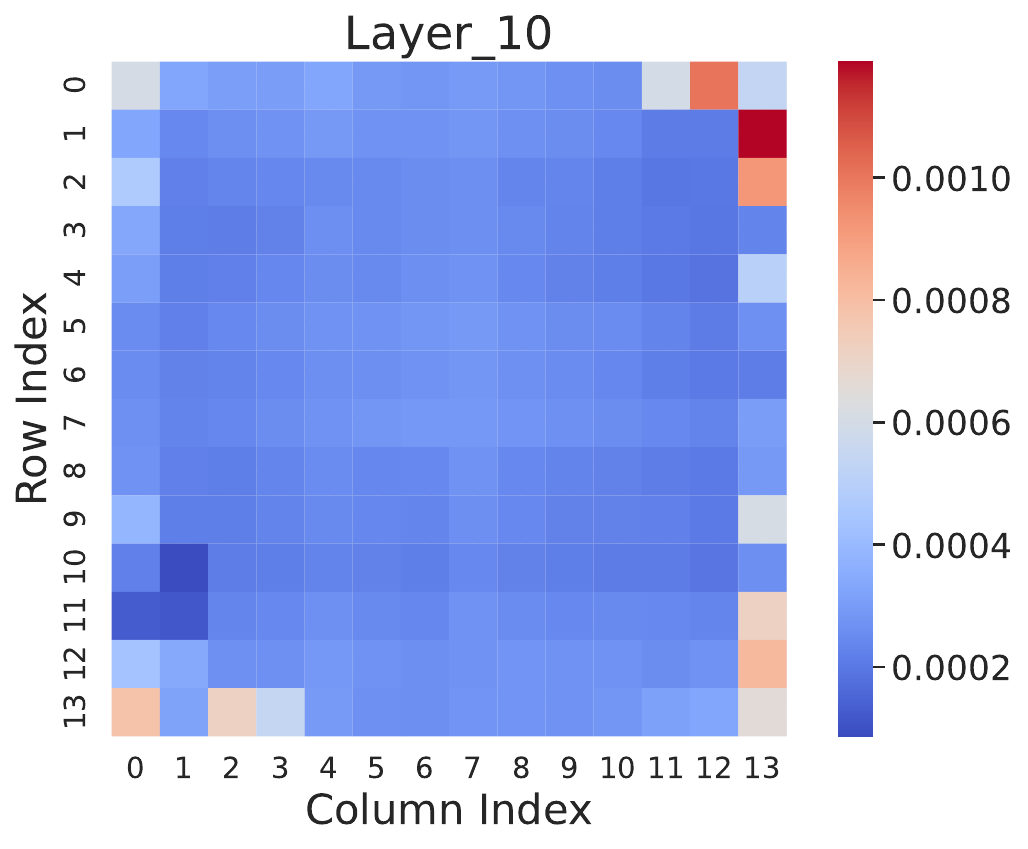}
  \end{subfigure}

  \caption{Visualization of local attention bias in Layer 1 and 10. Each block illustrates the sum of attention scores from prompt tokens to each spatial position across all frames.}
  \label{fig:position_heatmap_all}
\end{figure}

\textbf{\textit{Observation 2.}} \textbf{Local attention bias.} In shallow layers of language models, we observe that across all frames of videos, Video LLMs intrinsically allocate substantial attention to a few specific spatial positions, and remarkably, these spatial positions remain fixed regardless of variations in video and text content. In Fig.~\ref{fig:position_heatmap_all}, each grid shows the total attention received by that spatial position across all frames, and the 196 grids correspond to the 14×14 spatial patches extracted from each video frame. In layer 1, the spatial position receiving highest attention (Row 11, Column 1) receives 5.77 times the average attention of all visual tokens, clearly illustrating the presence of local bias.

Due to the biases stated above, directly leveraging raw attention scores as importance metric is not accurate and thus suboptimal. To achieve more precise pruning of visual tokens, we propose the method described in the next subsection. Based on the correction of the two types of biases mentioned above, we can effectively leverage attention scores to guide the token pruning process.

% two parts of attention bias, which can cause the sub-optimal choosing of representative tokens: 
% global attention bias and inside-frame attention bias
% global attention bias: attention scores are leaning towards the latter part of visual token sequences;
% inside-frame attention bias: most of attention scores are concentrated on a few positions;

% The above attention bias phenomenons collectively lead to bias in token selection, and the poor performance of attention-based selecting algorithms inside Language Model.

% \subsection{text guided segment-wise pruning}
% To correct global bias, we partition videos into multiple segments, and conduct token pruning process inside each segment. We also introduces text encoder to identify most text-related segments, and to keep different num of visual tokens for different segments. 

\subsection{Our pipeline: AdaTP}
\label{subsec:pipeline}

\begin{figure*}[htbp]
    \centering
    \includegraphics[width=\textwidth]{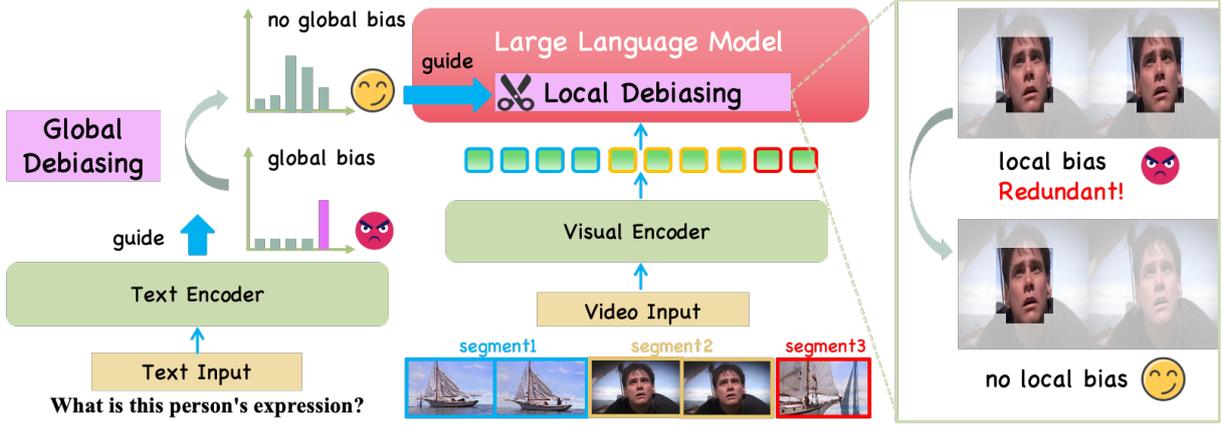}
    \caption{Illustration of our AdaTP pipeline. AdaTP mitigates attention bias by selectively retaining critical tokens based on text relevance and spatial diversity. The Global Debiasing Module identifies significant video segments, and the Local Debiasing Module further alleviates intra-segment spatial redundancy.}
    \label{fig:pipeline}
\end{figure*}

Based on the above observations, we propose a new token pruning pipeline named Attention-Debiased Token Pruning for Video LLMs (AdaTP). 

Due to temporal coherence of videos, a video can typically be divided into several parts, where frames within each part are highly similar, while different parts convey diverse content, thus we divide each video into multiple segments. For each video, its visual feature sequence $VE(v)=(v_1, v_2, \ldots, v_n)$ are partitioned into consecutive segments $\mathcal{S} = (s_1, s_2, \ldots, s_m)$ based on cosine similarity between adjacent frames, where $m$ denotes the number of segments. Given a threshold $\tau_s$, two adjacent frames $v_i$ and $v_{i+1}$ are assigned to the same segment if $cos\_sim(v_i, v_{i+1}) \geq \tau_s$. Otherwise, a new segment is started at frame $v_{i+1}$. 

Correspondingly, $X^l_v$ can be represented in segment-by-segment manner:

\begin{equation}
X^l_v=[X^l_{s_1}, X^l_{s_2}, \ldots, X^l_{s_m}].
\label{eq:segment_representation}
\end{equation}

Inspired by AIM~\cite{zhong2024aim} and PyramidDrop~\cite{xing2024pyramiddrop}, we adopt progressive token pruning paradigm. In shallow layers of the language model, we gradually prune a certain number of visual tokens at each layer.

As show in Fig.~\ref{fig:pipeline}, our pipeline mainly consists of two modules: Global Debiasing Module and Local Debiasing Module. The Global Debiasing Module leverages the text encoder to identify significant segments which are highly relevant to textual information, and assigns higher retention ratios to them. The local debiasing module operates within each segment, removing redundant visual tokens that occupy the same spatial positions across these adjacent and similar frames.

\begin{table*}[t]
\centering
\scriptsize
\caption{Comparison of our method with other baselines across various benchmarks on three Video LLMs. Within the same comparison, the highest compression ratio and highest performances are highlighted in bold.}
\label{tab:method-comparison}
\resizebox{\textwidth}{!}{
\begin{tabular}{@{}l l cccccccc@{}}
\toprule
\textbf{Model} & \textbf{Method} & \textbf{FLOPs} 
& \multicolumn{4}{c}{\textbf{VideoMME}} 
& \textbf{MLVU} & \textbf{LongVideoBench} 
& \textbf{Avg} \\
\cmidrule(lr){4-7}
\textbf{} & \textbf{} & \textbf{} 
& \textbf{short} & \textbf{medium} & \textbf{long} & \textbf{overall} 
& \textbf{} & \textbf{} & \textbf{} \\
\midrule

\multirow{12}{*}{\textbf{LLaVA-OneVision-0.5B}} 
 & vanilla              & 100.00\%       & 53.4 & 40.9 & 36.9 & 43.74 & 46.45 & 47.64 & 45.94 \\
\cmidrule{2-10}
 & FastV\textsuperscript{\textcolor{gray}{(ECCV'24)}}                & 55.19\%     & 53.1 & \textbf{42.1} & 36.6 & 43.93 & 45.07 & 45.77 & 44.92 \\
 & VisionZip\textsuperscript{\textcolor{gray}{(CVPR'25)}}            & 44.38\%     & 53.4 & 40.7 & 37.4 & 43.85 & 47.23 & 47.12 & 46.07 \\
 & Dycoke\textsuperscript{\textcolor{gray}{(CVPR'25)}}               & 50.32\%     & 55.0 & 40.4 & \textbf{38.7} & \textbf{44.70} & 47.24 & 46.97 & 46.30 \\
 & \textbf{AdaTP (ours)} & \textbf{41.08\%} & \textbf{55.4} & 41.8 & 36.4 & 44.56 & \textbf{47.36} & \textbf{47.79} & \textbf{46.57} \\
\cmidrule{2-10}
 & FastV                & 43.66\%     & 52.3 & \textbf{41.2} & 36.4 & 43.33 & 45.38 & 45.85 & 44.85 \\
 & VisionZip            & 39.37\%     & 53.8 & 41.0 & 37.0 & 43.93 & \textbf{47.54} & 46.15 & 45.87 \\
 & Dycoke               & 36.50\%         & 54.4  & 40.3  & \textbf{37.7}  & 44.10   & 45.26   & 46.37   & 45.24   \\
 & \textbf{AdaTP (ours)} & \textbf{33.99\%}  & \textbf{55.8} & \textbf{41.2} & 36.4 & \textbf{44.48} & 47.25 & \textbf{48.24} & \textbf{46.66} \\
\cmidrule{2-10}
 & FastV                & 32.13\%     & 48.0 & 41.4 & 36.7 & 42.04 & 44.47 & 45.03 & 43.85 \\
 & VisionZip            & 32.36\%     & 52.6 & 40.2 & 36.1 & 42.96 & 46.12 & 46.30 & 45.13 \\
 & Dycoke               & 32.86\%     & 53.9 & 39.2 & 36.1 & 43.07 & 42.23 & 44.50 & 43.27 \\
 & \textbf{AdaTP (ours)} & \textbf{26.43\%} & \textbf{54.7} & \textbf{42.7} & \textbf{37.6} & \textbf{44.96} & \textbf{47.58} & \textbf{47.05} & \textbf{46.53} \\
\midrule

\multirow{15}{*}{\textbf{LLaVA-OneVision-7B}} 
 & vanilla              & 100.00\%       & 70.0 & 56.7 & 48.7 & 58.44 & 63.25 & 56.32 & 59.34 \\
\cmidrule{2-10}
 & FastV                & 54.64\%     & 70.0 & 55.0 & 47.2 & 57.41 & 62.22 & 55.65 & 58.43 \\
 & VisionZip            & 46.38\%     & \textbf{70.9} & \textbf{56.9} & 49.0 & \textbf{58.93} & 63.22 & 55.65 & 59.27 \\
 & Dycoke               & 50.32\%     & 70.1 & 54.2 & 48.7 & 57.67 & 63.16 & 56.39 & 59.39 \\
 & \textbf{AdaTP (ours)} & \textbf{43.00\%} & 70.4 & 56.2 & \textbf{49.4} & 58.70 & \textbf{63.95} & \textbf{56.69} & \textbf{59.78} \\
\cmidrule{2-10}
 & FastV                & 43.18\%     & 68.6 & 55.1 & 47.9 & 57.19 & 61.35 & 55.12 & 57.89 \\
 & VisionZip            & 39.37\%     & \textbf{71.1} & 55.8 & \textbf{49.7} & 58.85 & 63.09 & 56.02 & 59.32 \\
 & Dycoke               & 36.44\%     & 67.6  & 53.3  & 48.0  & 56.30   & 61.13   & \textbf{56.69}   & 58.04   \\
 & \textbf{AdaTP (ours)} & \textbf{35.48\%}     & 71.0 & \textbf{56.8} & 48.9 & \textbf{58.89} & \textbf{64.27} & 55.95 & \textbf{59.70} \\
\cmidrule{2-10}
 & FastV                & 35.20\%     & 66.1 & 53.6 & 48.1 & 55.93 & 60.82 & 51.91 & 56.22 \\
 & VisionZip            & 27.95\%     & 69.2 & \textbf{57.7} & \textbf{49.2} & 58.70 & 63.13 & \textbf{56.17} & 59.33 \\
 & Dycoke               & 32.84\%     & 67.1 & 53.1 & 46.6 & 55.59 & 60.90 & 55.95 & 57.48 \\
 & \textbf{AdaTP (ours)} & \textbf{27.30\%} & \textbf{71.0} & 56.9 & \textbf{49.2} & \textbf{59.04} & \textbf{63.47} & 56.02 & \textbf{59.51} \\
\midrule

\multirow{15}{*}{\textbf{LLaVA-Video-7B}} 
 & vanilla              & 100.00\%       & 70.0 & 56.7 & 48.7 & 58.44 & 63.25 & 56.32 & 59.34 \\
\cmidrule{2-10}
 & FastV                & 54.89\%     & 69.4 & 59.2 & \textbf{50.2} & 59.63 & 61.56 & 54.15 & 58.45 \\
 & VisionZip            & 44.42\%     & 72.6 & 58.9 & 48.3 & 59.93 & 61.55 & 55.72 & 59.07 \\
 & Dycoke               & 59.26\%     & 72.2 & 57.7 & 47.8 & 59.22 & 60.67 & 55.57 & 58.49 \\
 & \textbf{AdaTP (ours)} & \textbf{44.26\%} & \textbf{72.7} & \textbf{59.9} & 49.9 & \textbf{60.81} & \textbf{62.10} & \textbf{56.32} & \textbf{59.74} \\
\cmidrule{2-10}
 & FastV                & 43.61\%     & 67.2 & 57.1 & 47.9 & 57.41 & 60.00 & 52.81 & 56.74 \\
 & VisionZip            & 37.52\%     & 71.0 & 58.1 & 49.3 & 59.48 & 61.34 & 54.97 & 58.60 \\
 & Dycoke               & 38.71\%     & 68.9  & 56.7  & 47.8  & 57.78   & 58.83   & 54.23   & 56.95   \\
 & \textbf{AdaTP (ours)} & \textbf{36.63\%}     & \textbf{72.2} & \textbf{59.8} & \textbf{50.0} & \textbf{60.67} & \textbf{61.42} & \textbf{56.77} & \textbf{59.62} \\
\cmidrule{2-10}
 & FastV                & 32.33\%     & 62.6 & 54.1 & 48.0 & 54.89 & 57.80 & 51.09 & 54.59 \\
 & VisionZip            & 28.69\%     & \textbf{70.4} & 57.8 & 49.3 & 59.19 & 59.81 & 54.08 & 57.69 \\
 & Dycoke               & 35.36\%     & 68.1 & 55.7 & 46.2 & 56.67 & 59.02 & 53.78 & 56.49 \\
 & \textbf{AdaTP (ours)} & \textbf{28.27\%}     & 70.3 & \textbf{59.1} & \textbf{51.7} & \textbf{60.37} & \textbf{60.95} & \textbf{55.72} & \textbf{59.01} \\
\bottomrule

\end{tabular}
}
\end{table*}

(1) \textbf{Global Debiasing Module.} We introduce the text encoder aligned with the visual encoder in Video LLMs to tackle global attention bias. For $i$-th video frame, the visual encoder produces a global visual token $v_{pi} \in \mathbb{R}^d$. Meanwhile, the input text is processed by the text encoder $TE(\cdot)$ to obtain a global textual token $t_p \in \mathbb{R}^d$. Text relevance of each frame can then be measured by computing the cosine similarity between $v_{pi}$ and $t_p$:
\begin{equation}
\label{eq:sim}
sim_i = \frac{v_{pi}^\top t_p}{\|v_{pi}\| \cdot \|t_p\|}, i=1,2,\ldots, n,
\end{equation}
where $\|\cdot\|$ denotes the L2 norm. We estimate the text relevancy of each segment by averaging similarities of all frames within it. We then introduce a threshold parameter $\tau_t$, and segments with similarity exceeding $\tau_t$ are classified as highly relevant to the text (hereafter referred to as significant segments for convenience). The selected significant segments can be represented as set $\mathcal{S'}\subseteq \mathcal{S}$.

\begin{table*}[t]
\centering
\caption{Ablation study on different modules. Seg. means segment-aware token pruning, Global.D denotes Global Debiasing Module, while Local.D denotes Local Debiasing Module. \cmark\ indicates the bias is corrected; and \xmark\ indicates it is not.}
\label{tab:ablation}
\scriptsize
\resizebox{\linewidth}{!}{
\begin{tabular}{@{}ccccccccc@{}}
\toprule
\textbf{Seg.} & \textbf{Global.D} & \textbf{Local.D} & \textbf{FLOPs} & \textbf{VideoMME} & \textbf{MVLU} & \textbf{LongVideoBench} & \textbf{Avg} \\
\midrule
\xmark & \xmark & \xmark & 27.65\% & 43.67          & 46.58          & 44.88          & 45.04          \\
\cmark & \xmark & \xmark & 27.76\% & 44.63 & 46.26 & 46.67 & 45.85 \\
\cmark & \cmark & \xmark & 27.33\% & 44.63 & 47.24 & \textbf{47.19} & 46.35 \\
\cmark & \cmark & \cmark & \textbf{26.43\%} & \textbf{44.96} & \textbf{47.58} & 47.05 & \textbf{46.53} \\
\bottomrule
\end{tabular}
}
\end{table*}

We perform token pruning within each segment, assigning higher retention ratios to significant ones. To be specific, we define an overall compression ratio $r$, retaining totally $L_v^l \cdot r$ visual tokens, distributed across segments. For each segment $s$, the number of retained tokens is proportional to its frame count $\|s\|$. For significant segments $s_i \in \mathcal{S'}$, this allocation is upscaled by a factor $\alpha_{\text{boost}}$, but constrained such that the total retained tokens in all significant segments do not exceed $\gamma_{\text{cap}} \cdot L_v^l$, ensuring coherence and preventing over-concentration. This strategy effectively preserves critical content while reducing redundancy.

The retention ratio of significant segments are:
\begin{equation}
r_1 = \min \left( 
    \alpha_{\text{boost}},\ 
    \frac{n}{\sum_{s\in \mathcal{S'}}\|s\|} \cdot \gamma_{\text{cap}}
\right) \cdot r,
\label{eq00}
\end{equation}
while the retention ratio of other segments are:
\begin{equation}
r_2 = \frac{r \cdot n - r_1 \cdot \sum_{s\in \mathcal{S'}}\|s\|}{\sum_{s\notin \mathcal{S'}}\|s\|}.
\label{eq01}
\end{equation}

Within each segment $s_i$, we select the above ratio of visual tokens ($r_1$ for significant segments, and $r_2$ for others) with the highest attention scores.

\begin{algorithm}[htbp]
\caption{Local Debiasing Module}
\label{alg:local_dedup}
\KwIn{
  $V = \{v_1, v_2, \dots, v_n\}$: selected visual tokens within a segment. \\
  $\text{pos}(v_i)$: spatial position of token $v_i$; \\
  $\text{score}(v_i)$: attention score of token $v_i$.
}
\KwOut{
  $V' \subseteq V$: deduplicated token set.
}
Sort $V$ in descending order by $\text{score}(v_i)$\;
$P \leftarrow \emptyset$ \tcp*{Used spatial positions}
$V' \leftarrow \emptyset$ \tcp*{Selected tokens}
\For{$v_i$ in $V$}{
  \If{$\text{pos}(v_i) \notin P$}{
    Add $v_i$ to $V'$\;
    Add $\text{pos}(v_i)$ to $P$\;
  }
}
\Return $V'$
\end{algorithm}

(2) \textbf{Local Debiasing Module.} 
Due to the similarity among frames within the same segment, visual tokens from identical spatial positions among the selected high-attention tokens are redundant. Our Local Debiasing Module removes spatially redundant visual tokens, thereby further reducing unnecessary computational overhead.

We propose a simple yet effective strategy to address this redundancy. Specifically, after obtaining the set of visual tokens with the highest attention scores within each segment, we sort them in descending order based on their scores. We then iterate through the sorted list, and for each token, we check whether another token from the same spatial position in other frames has already been selected. If so, the current token is considered redundant and discarded; otherwise, it is retained. This ensures that only one representative token is kept for each spatial location within the segment. The full procedure is detailed in Algorithm~\ref{alg:local_dedup}.

Moreover, with this design, our pipeline fully accounts for diverse and unique characteristics of each video by adaptively adjusting the visual token compression rate. Rather than applying a fixed pruning scheme, more dynamic videos—those exhibiting frequent scene changes or significant motion—are typically divided into more segments, allowing them to retain a larger proportion of visual tokens. In contrast, static videos with minimal variation across frames are partitioned into fewer segments and accordingly allocated fewer tokens. This ensures that token pruning is more aligned with the content dynamics of the videos.

\section{Experiments}

\subsection{Implementation Details}
\begin{figure*}[htbp]
    \centering
    \includegraphics[width=\linewidth, keepaspectratio]{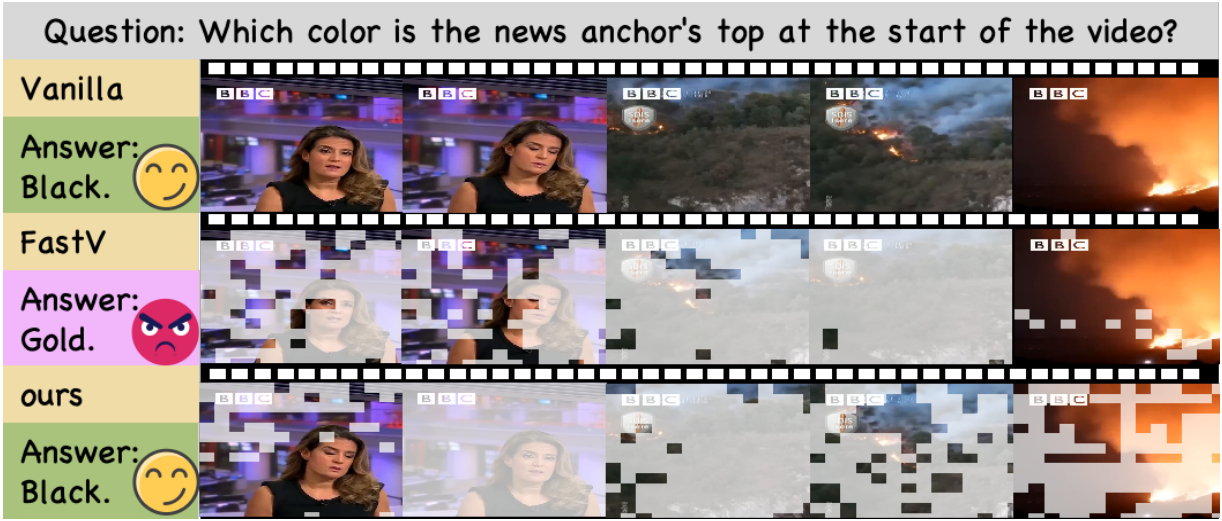}
    \caption{Visualization example sampled from the VideoMME dataset. We adopt attention scores from layer 1 to generate the visualization results, in which we perform significant token pruning.}
    \label{fig:vis383}
\end{figure*}
% \begin{figure*}[htbp]
%     \centering
%     \begin{subfigure}[b]{0.45\textwidth}
%         \includegraphics[width=\linewidth]{latex/figures/tau_t.pdf}
%         \caption{$\tau_t$}
%     \end{subfigure}
%     \hfill
%     \begin{subfigure}[b]{0.45\textwidth}
%         \includegraphics[width=\linewidth]{latex/figures/alpha_{\text{boost}}.pdf}
%         \caption{$\alpha_{\text{boost}}$}
%     \end{subfigure}

%     \begin{subfigure}[b]{0.45\textwidth}
%         \includegraphics[width=\linewidth]{latex/figures/gamma_{\text{cap}}.pdf}
%         \caption{$\gamma_{\text{cap}}$}
%     \end{subfigure}
%     \hfill
%     \begin{subfigure}[b]{0.45\textwidth}
%         \includegraphics[width=\linewidth]{latex/figures/tau_s.pdf}
%         \caption{$\tau_s$}
%     \end{subfigure}
    
%     \caption{Ablation study on hyperparameters $\tau_t$, $\alpha_{\text{boost}}$, $\gamma_{\text{cap}}$, and $\tau_s$.}
%     \label{fig:ablation:hyperparameters}
% \end{figure*}

\textbf{Benchmarks.} We evaluate the performance of Video LLMs on three widely used video understanding benchmarks: VideoMME~\cite{fu2024video}, MLVU~\cite{zhou2024mlvu}, and LongVideoBench~\cite{wu2024longvideobench}. The VideoMME dataset is divided into short (1-3 minutes), medium (3-30 minutes), and long subsets (30-60 minutes), each containing 900 video-question pairs. The MLVU dataset consists of videos ranging from 3 minutes to 2 hours in length and includes nine distinct evaluation tasks, totaling 2,174 samples. LongVideoBench comprises video-question pairs across 17 categories, ranging from 1 to 60 minutes; we use its validation set, which contains 1,337 samples. These datasets cover various types, durations, and scenarios of videos, providing a fair and comprehensive evaluation for video understanding ability of Video LLMs.

\textbf{Model Settings.} We apply our method on three types out of two representative Video LLM architectures: LLaVA-OneVision-0.5B, LLaVA-OneVision-7B,~\cite{Li2024LLaVAOneVision} and LLaVA-Video-7B~\cite{lin2023video}. Under our settings, LLaVA-OneVision models samples 32 frames, generates 196 visual tokens per frame; LLaVA-Video model samples 20 frames, generates 182 visual tokens per frame, including 13 added newline tokens. All experiments are conducted on a single NVIDIA 3090 GPU. All FLOPs statistics are collected using the torch.profiler package provided by PyTorch for consistent and accurate measurement.

Our progressive pruning process starts from the second layer of the language model until the $(N-12)$-th layer, where $N$ denotes the total number of layers in language model ($N=24$ for LLaVA-OneVision-0.5B model, and $N=28$ for LLaVA-OneVision-7B and LLaVA-Video-7B models). Before passing to the second layer, we retain $p\cdot 40\%$ of the visual tokens; after layer $N-12$, only $p \cdot 12\%$ of the tokens are kept; for all intermediate layers, tokens are removed at a uniform rate. Here $p$ is the parameter for us to adjust the compression ratio. For hyperparameters of our pipeline modules, we set $\tau_t=0.05,\alpha_{\text{boost}}=2.2, \gamma_{\text{cap}}=0.75, \tau_s=0.95$ for all models. For three different compression ratios, we set $p$ to 2.0, 1.5 and 1.0, respectively.

\textbf{Baselines.} We adopt three training-free visual token compression methods: FastV, VisionZip and Dycoke. We use official codes of these methods for fair comparison, and conduct all evaluations using LMMs-Eval~\cite{zhang2024lmms}. Detailed settings please refer to Appendix~\ref{subsec:experimantal_settings}.

\subsection{Main Results}

Tab.\ref{tab:method-comparison} compares our AdaTP with state-of-the-art methods on aforementioned models. On all three models and with three compression ratios, our AdaTP outperforms state-of-the-art baselines. On both LLaVA-OneVision-0.5B and LLaVA-OneVision-7B, our method surprisingly surpasses the vanilla model. We attribute this improvement to our pipeline’s ability to eliminate redundant visual information and guide the model to focus more on critical tokens. On LLaVA-Video-7B, our AdaTP also showed clear advantage over other baselines. 

As the compression ratio increases, all other methods experience a significant performance drop, whereas our method consistently maintains strong performance. This shows the robustness and effectiveness of our AdaTP.

\subsection{Ablation Studies}
We conduct ablation studies on modules in our pipeline to prove their functionality, and also on hyperparameters to show its robustness. To ensure consistency and clarity in our presentation, we conduct the ablation study on LLaVA-onevision-0.5B, and $p$ is set to 1.

\textbf{Ablation study on debiasing modules.} We first conduct ablation studies on our two debiasing modules and our segment-aware design. Without Local Debiasing module, intra-segment spatial redundancy will be retained; without the Global Debiasing Module, significant segments will not be distinguished. The results demonstrate the effectiveness of segment-aware design and both of our two debiasing modules, further highlight the impact of attention bias on the performance of pruned Video LLMs. The results are shown in Tab.~\ref{tab:ablation}.

\textbf{Ablation study on hyperparameters.} We evaluate the performance of hyperparameters $\tau_s, \tau_t, \alpha_{\text{boost}}, \ \text{and} \  \gamma_{\text{cap}}$. In this part, we also set $p$ to 1. The results are shown in Appendix~\ref{subsec:ablation-hyperparameters}.

For progressive layer-by-layer token pruning, we start pruning from Layer 2 while keeping the last 12 layers unpruned across all Video LLMs. A uniform pruning rate is applied between the pruned layers, with a fixed number of tokens discarded after each layer. In LLaVA-OneVision-0.5B~\cite{Li2024LLaVAOneVision}, which consists of 24 layers, pruning depth is by default set to Layer 2-12. Variants with different pruning depths are also presented in Tab.~\ref{tab:ablation-depth} for comparison. 

\begin{table}[t]
\captionsetup{font=small, skip=4pt}
\centering
\caption{Ablation study on pruning layer depths.}
\label{tab:ablation-depth}
\resizebox{\linewidth}{!}{
\begin{tabular}{cccccc}
\toprule
\textbf{layers} & \textbf{FLOPs} & \textbf{VideoMME} & \textbf{MLVU} & \textbf{LongVideoBench} & \textbf{Avg} \\
\midrule
2-8 & 24.46\% & 44.30 & 47.12 & 46.30 & 45.91 \\
2-10 & 25.45\% & 44.74 & 47.90 & 46.60 & 46.41 \\
2-12 & 26.43\% & 44.96 & 47.58 & 47.05 & 46.53 \\
2-14 & 27.37\% & 44.52 & 46.89 & 47.57 & 46.33 \\
2-16 & 28.40\% & 44.78 & 47.33 & 47.57 & 46.56 \\
\bottomrule
\end{tabular}
}
\end{table}

\subsection{Visualization}

To further illustrate the effectiveness of our method, we visualize the token pruning process. The pruned tokens are visually masked with a white overlay to indicate their exclusion. As shown in Fig.~\ref{fig:vis383}, for this video-question pair, the vanilla model produces the correct answer, while FastV gives an incorrect response. In contrast, our method retains significant visual tokens and answers correctly. It can be observed that due to the influence of global bias, FastV fails to extract text-relevant content effectively and retains a large number of visual tokens from the latter part of the sequence. Our approach, however, accurately identifies the video segments closely aligned with the textual semantics, \textit{i.e.} the first two frames containing the news anchor. Among these two visually similar frames, our method retains informative content from only one frame, thereby effectively addressing the visual redundancy caused by local bias.

\section{Conclusion}
In this paper, we identified and thoroughly analyzed the attention bias phenomenon in Video LLMs, and proposed AdaTP, a training-free token pruning pipeline. Our AdaTP effectively addressed the issue of attention bias, relieved the heavy computational burden of Video LLMs while retaining critical visual contents. Extensive experiments on different Video LLMs and video understanding benchmarks have demonstrated the effectiveness and robustness of our method.

\section*{Limitations}

Due to computational resource constraints, our proposed method has not yet been validated on larger-scale Video LLMs. In addition, our pipeline introduces a relatively large number of hyperparameters, which may increase its complexity. However, we have conducted comprehensive ablation studies on them, demonstrating that the method remains robust across these hyperparameters.

% \section*{Acknowledgments}

% This document has been adapted
% by Steven Bethard, Ryan Cotterell and Rui Yan
% from the instructions for earlier ACL and NAACL proceedings, including those for
% ACL 2019 by Douwe Kiela and Ivan Vuli\'{c},
% NAACL 2019 by Stephanie Lukin and Alla Roskovskaya,
% ACL 2018 by Shay Cohen, Kevin Gimpel, and Wei Lu,
% NAACL 2018 by Margaret Mitchell and Stephanie Lukin,
% Bib\TeX{} suggestions for (NA)ACL 2017/2018 from Jason Eisner,
% ACL 2017 by Dan Gildea and Min-Yen Kan,
% NAACL 2017 by Margaret Mitchell,
% ACL 2012 by Maggie Li and Michael White,
% ACL 2010 by Jing-Shin Chang and Philipp Koehn,
% ACL 2008 by Johanna D. Moore, Simone Teufel, James Allan, and Sadaoki Furui,
% ACL 2005 by Hwee Tou Ng and Kemal Oflazer,
% ACL 2002 by Eugene Charniak and Dekang Lin,
% and earlier ACL and EACL formats written by several people, including
% John Chen, Henry S. Thompson and Donald Walker.
% Additional elements were taken from the formatting instructions of the \emph{International Joint Conference on Artificial Intelligence} and the \emph{Conference on Computer Vision and Pattern Recognition}.

% Bibliography entries for the entire Anthology, followed by custom entries
%\bibliography{anthology,custom}
% Custom bibliography entries only
\bibliography{custom}

\begin{thebibliography}{37}
\providecommand{\natexlab}[1]{#1}

\bibitem[{Alayrac et~al.(2022)Alayrac, Donahue, Luc, Miech, Barr, Hasson, Lenc, Mensch, Millican, Reynolds et~al.}]{alayrac2022flamingo}
Jean-Baptiste Alayrac, Jeff Donahue, Pauline Luc, Antoine Miech, Iain Barr, Yana Hasson, Karel Lenc, Arthur Mensch, Katherine Millican, Malcolm Reynolds, and 1 others. 2022.
\newblock Flamingo: a visual language model for few-shot learning.
\newblock \emph{Advances in neural information processing systems}, 35:23716--23736.

\bibitem[{Chen et~al.(2024{\natexlab{a}})Chen, Zhao, Liu, Bai, Lin, Zhou, and Chang}]{chen2024image}
Liang Chen, Haozhe Zhao, Tianyu Liu, Shuai Bai, Junyang Lin, Chang Zhou, and Baobao Chang. 2024{\natexlab{a}}.
\newblock An image is worth 1/2 tokens after layer 2: Plug-and-play inference acceleration for large vision-language models.
\newblock In \emph{European Conference on Computer Vision}, pages 19--35. Springer.

\bibitem[{Chen et~al.(2024{\natexlab{b}})Chen, Wang, Cao, Liu, Gao, Cui, Zhu, Ye, Tian, Liu et~al.}]{chen2024expanding}
Zhe Chen, Weiyun Wang, Yue Cao, Yangzhou Liu, Zhangwei Gao, Erfei Cui, Jinguo Zhu, Shenglong Ye, Hao Tian, Zhaoyang Liu, and 1 others. 2024{\natexlab{b}}.
\newblock Expanding performance boundaries of open-source multimodal models with model, data, and test-time scaling.
\newblock \emph{arXiv preprint arXiv:2412.05271}.

\bibitem[{Fu et~al.(2024)Fu, Dai, Luo, Li, Ren, Zhang, Wang, Zhou, Shen, Zhang et~al.}]{fu2024video}
Chaoyou Fu, Yuhan Dai, Yongdong Luo, Lei Li, Shuhuai Ren, Renrui Zhang, Zihan Wang, Chenyu Zhou, Yunhang Shen, Mengdan Zhang, and 1 others. 2024.
\newblock Video-mme: The first-ever comprehensive evaluation benchmark of multi-modal llms in video analysis.
\newblock \emph{arXiv preprint arXiv:2405.21075}.

\bibitem[{Li et~al.(2024)Li, Zhang, Guo, Zhang, Li, Zhang, Zhang, Li, Liu, and Li}]{Li2024LLaVAOneVision}
Bo~Li, Yuanhan Zhang, Dong Guo, Renrui Zhang, Feng Li, Hao Zhang, Kaichen Zhang, Yanwei Li, Ziwei Liu, and Chunyuan Li. 2024.
\newblock \href {https://arxiv.org/abs/2408.03326} {Llava-onevision: Easy visual task transfer}.
\newblock \emph{arXiv preprint arXiv:2408.03326}.

\bibitem[{Li et~al.(2023)Li, Li, Savarese, and Hoi}]{li2023blip}
Junnan Li, Dongxu Li, Silvio Savarese, and Steven Hoi. 2023.
\newblock Blip-2: Bootstrapping language-image pre-training with frozen image encoders and large language models.
\newblock In \emph{International conference on machine learning}, pages 19730--19742. PMLR.

\bibitem[{Liang et~al.(2022)Liang, Ge, Tong, Song, Wang, and Xie}]{liang2022not}
Youwei Liang, Chongjian Ge, Zhan Tong, Yibing Song, Jue Wang, and Pengtao Xie. 2022.
\newblock Not all patches are what you need: Expediting vision transformers via token reorganizations.
\newblock \emph{arXiv preprint arXiv:2202.07800}.

\bibitem[{Lin et~al.(2023)Lin, Ye, Zhu, Cui, Ning, Jin, and Yuan}]{lin2023video}
Bin Lin, Yang Ye, Bin Zhu, Jiaxi Cui, Munan Ning, Peng Jin, and Li~Yuan. 2023.
\newblock Video-llava: Learning united visual representation by alignment before projection.
\newblock \emph{arXiv preprint arXiv:2311.10122}.

\bibitem[{Liu et~al.(2024{\natexlab{a}})Liu, Li, Li, and Lee}]{liu2024improved}
Haotian Liu, Chunyuan Li, Yuheng Li, and Yong~Jae Lee. 2024{\natexlab{a}}.
\newblock Improved baselines with visual instruction tuning.
\newblock In \emph{Proceedings of the IEEE/CVF Conference on Computer Vision and Pattern Recognition}, pages 26296--26306.

\bibitem[{Liu et~al.(2024{\natexlab{b}})Liu, Li, Li, Li, Zhang, Shen, and Lee}]{liu2024llavanext}
Haotian Liu, Chunyuan Li, Yuheng Li, Bo~Li, Yuanhan Zhang, Sheng Shen, and Yong~Jae Lee. 2024{\natexlab{b}}.
\newblock Llavanext: Improved reasoning, ocr, and world knowledge.

\bibitem[{Liu et~al.(2023)Liu, Li, Wu, and Lee}]{liu2023visual}
Haotian Liu, Chunyuan Li, Qingyang Wu, and Yong~Jae Lee. 2023.
\newblock Visual instruction tuning.
\newblock \emph{Advances in neural information processing systems}, 36:34892--34916.

\bibitem[{Maaz et~al.(2023)Maaz, Rasheed, Khan, and Khan}]{maaz2023video}
Muhammad Maaz, Hanoona Rasheed, Salman Khan, and Fahad~Shahbaz Khan. 2023.
\newblock Video-chatgpt: Towards detailed video understanding via large vision and language models.
\newblock \emph{arXiv preprint arXiv:2306.05424}.

\bibitem[{Marin et~al.(2021)Marin, Chang, Ranjan, Prabhu, Rastegari, and Tuzel}]{marin2021token}
Dmitrii Marin, Jen-Hao~Rick Chang, Anurag Ranjan, Anish Prabhu, Mohammad Rastegari, and Oncel Tuzel. 2021.
\newblock Token pooling in vision transformers.
\newblock \emph{arXiv preprint arXiv:2110.03860}.

\bibitem[{Radford et~al.(2021)Radford, Kim, Hallacy, Ramesh, Goh, Agarwal, Sastry, Askell, Mishkin, Clark, Krueger, and Sutskever}]{Radford2021LearningTV}
Alec Radford, Jong~Wook Kim, Chris Hallacy, Aditya Ramesh, Gabriel Goh, Sandhini Agarwal, Girish Sastry, Amanda Askell, Pamela Mishkin, Jack Clark, Gretchen Krueger, and Ilya Sutskever. 2021.
\newblock \href {https://api.semanticscholar.org/CorpusID:231591445} {Learning transferable visual models from natural language supervision}.
\newblock In \emph{International Conference on Machine Learning}.

\bibitem[{Rao et~al.(2021)Rao, Zhao, Liu, Lu, Zhou, and Hsieh}]{rao2021dynamicvit}
Yongming Rao, Wenliang Zhao, Benlin Liu, Jiwen Lu, Jie Zhou, and Cho-Jui Hsieh. 2021.
\newblock Dynamicvit: Efficient vision transformers with dynamic token sparsification.
\newblock \emph{Advances in neural information processing systems}, 34:13937--13949.

\bibitem[{Shang et~al.(2024)Shang, Cai, Xu, Lee, and Yan}]{shang2024llava}
Yuzhang Shang, Mu~Cai, Bingxin Xu, Yong~Jae Lee, and Yan Yan. 2024.
\newblock Llava-prumerge: Adaptive token reduction for efficient large multimodal models.
\newblock \emph{arXiv preprint arXiv:2403.15388}.

\bibitem[{Shen et~al.(2025{\natexlab{a}})Shen, Gong, He, Zhang, Liu, Zhao, and Ding}]{shen2025fastvid}
Leqi Shen, Guoqiang Gong, Tao He, Yifeng Zhang, Pengzhang Liu, Sicheng Zhao, and Guiguang Ding. 2025{\natexlab{a}}.
\newblock Fastvid: Dynamic density pruning for fast video large language models.
\newblock \emph{arXiv preprint arXiv:2503.11187}.

\bibitem[{Shen et~al.(2024)Shen, Hao, He, Zhao, Zhang, Liu, Bao, and Ding}]{shen2024tempme}
Leqi Shen, Tianxiang Hao, Tao He, Sicheng Zhao, Yifeng Zhang, Pengzhang Liu, Yongjun Bao, and Guiguang Ding. 2024.
\newblock Tempme: Video temporal token merging for efficient text-video retrieval.
\newblock \emph{arXiv preprint arXiv:2409.01156}.

\bibitem[{Shen et~al.(2025{\natexlab{b}})Shen, He, Gong, Yang, Zhang, Liu, Zhao, and Ding}]{shen2025llava}
Leqi Shen, Tao He, Guoqiang Gong, Fan Yang, Yifeng Zhang, Pengzhang Liu, Sicheng Zhao, and Guiguang Ding. 2025{\natexlab{b}}.
\newblock Llava-mlb: Mitigating and leveraging attention bias for training-free video llms.
\newblock \emph{arXiv preprint arXiv:2503.11205}.

\bibitem[{Tao et~al.(2024)Tao, Qin, You, Sui, and Wang}]{tao2024dycoke}
Keda Tao, Can Qin, Haoxuan You, Yang Sui, and Huan Wang. 2024.
\newblock Dycoke: Dynamic compression of tokens for fast video large language models.
\newblock \emph{arXiv preprint arXiv:2411.15024}.

\bibitem[{Team et~al.(2023)Team, Anil, Borgeaud, Alayrac, Yu, Soricut, Schalkwyk, Dai, Hauth, Millican et~al.}]{team2023gemini}
Gemini Team, Rohan Anil, Sebastian Borgeaud, Jean-Baptiste Alayrac, Jiahui Yu, Radu Soricut, Johan Schalkwyk, Andrew~M Dai, Anja Hauth, Katie Millican, and 1 others. 2023.
\newblock Gemini: a family of highly capable multimodal models.
\newblock \emph{arXiv preprint arXiv:2312.11805}.

\bibitem[{Wang et~al.(2024{\natexlab{a}})Wang, Sun, Chen, Lin, Han, and Ding}]{wang2024cls}
Ao~Wang, Fengyuan Sun, Hui Chen, Zijia Lin, Jungong Han, and Guiguang Ding. 2024{\natexlab{a}}.
\newblock [cls] token tells everything needed for training-free efficient mllms.
\newblock \emph{arXiv preprint arXiv:2412.05819}.

\bibitem[{Wang et~al.(2024{\natexlab{b}})Wang, Bai, Tan, Wang, Fan, Bai, Chen, Liu, Wang, Ge et~al.}]{wang2024qwen2}
Peng Wang, Shuai Bai, Sinan Tan, Shijie Wang, Zhihao Fan, Jinze Bai, Keqin Chen, Xuejing Liu, Jialin Wang, Wenbin Ge, and 1 others. 2024{\natexlab{b}}.
\newblock Qwen2-vl: Enhancing vision-language model's perception of the world at any resolution.
\newblock \emph{arXiv preprint arXiv:2409.12191}.

\bibitem[{Wu et~al.(2024)Wu, Li, Chen, and Li}]{wu2024longvideobench}
Haoning Wu, Dongxu Li, Bei Chen, and Junnan Li. 2024.
\newblock Longvideobench: A benchmark for long-context interleaved video-language understanding.
\newblock \emph{Advances in Neural Information Processing Systems}, 37:28828--28857.

\bibitem[{Xiao et~al.(2023)Xiao, Tian, Chen, Han, and Lewis}]{xiao2023efficient}
Guangxuan Xiao, Yuandong Tian, Beidi Chen, Song Han, and Mike Lewis. 2023.
\newblock Efficient streaming language models with attention sinks.
\newblock \emph{arXiv preprint arXiv:2309.17453}.

\bibitem[{Xing et~al.(2024)Xing, Huang, Dong, Lu, Zhang, Zang, Cao, He, Wang, Wu et~al.}]{xing2024pyramiddrop}
Long Xing, Qidong Huang, Xiaoyi Dong, Jiajie Lu, Pan Zhang, Yuhang Zang, Yuhang Cao, Conghui He, Jiaqi Wang, Feng Wu, and 1 others. 2024.
\newblock Pyramiddrop: Accelerating your large vision-language models via pyramid visual redundancy reduction.
\newblock \emph{arXiv preprint arXiv:2410.17247}.

\bibitem[{Xu et~al.(2024)Xu, Zhao, Zhou, Lin, Ng, and Feng}]{xu2024pllava}
Lin Xu, Yilin Zhao, Daquan Zhou, Zhijie Lin, See~Kiong Ng, and Jiashi Feng. 2024.
\newblock Pllava: Parameter-free llava extension from images to videos for video dense captioning.
\newblock \emph{arXiv preprint arXiv:2404.16994}.

\bibitem[{Yang et~al.(2024)Yang, Chen, Tian, Wang, Li, Yu, and Jia}]{yang2024visionzip}
Senqiao Yang, Yukang Chen, Zhuotao Tian, Chengyao Wang, Jingyao Li, Bei Yu, and Jiaya Jia. 2024.
\newblock Visionzip: Longer is better but not necessary in vision language models.
\newblock \emph{arXiv preprint arXiv:2412.04467}.

\bibitem[{Yin et~al.(2022)Yin, Vahdat, Alvarez, Mallya, Kautz, and Molchanov}]{yin2022vit}
Hongxu Yin, Arash Vahdat, Jose~M Alvarez, Arun Mallya, Jan Kautz, and Pavlo Molchanov. 2022.
\newblock A-vit: Adaptive tokens for efficient vision transformer.
\newblock In \emph{Proceedings of the IEEE/CVF conference on computer vision and pattern recognition}, pages 10809--10818.

\bibitem[{Zhai et~al.(2023)Zhai, Mustafa, Kolesnikov, and Beyer}]{zhai2023sigmoid}
Xiaohua Zhai, Basil Mustafa, Alexander Kolesnikov, and Lucas Beyer. 2023.
\newblock Sigmoid loss for language image pre-training.
\newblock In \emph{Proceedings of the IEEE/CVF international conference on computer vision}, pages 11975--11986.

\bibitem[{Zhang et~al.(2024{\natexlab{a}})Zhang, Li, Zhang, Pu, Cahyono, Hu, Liu, Zhang, Yang, Li et~al.}]{zhang2024lmms}
Kaichen Zhang, Bo~Li, Peiyuan Zhang, Fanyi Pu, Joshua~Adrian Cahyono, Kairui Hu, Shuai Liu, Yuanhan Zhang, Jingkang Yang, Chunyuan Li, and 1 others. 2024{\natexlab{a}}.
\newblock Lmms-eval: Reality check on the evaluation of large multimodal models.
\newblock \emph{arXiv preprint arXiv:2407.12772}.

\bibitem[{Zhang et~al.(2024{\natexlab{b}})Zhang, Cheng, Lu, Zhuo, Wang, Cao, Guo, She, and Zhang}]{zhang2024cls}
Qizhe Zhang, Aosong Cheng, Ming Lu, Zhiyong Zhuo, Minqi Wang, Jiajun Cao, Shaobo Guo, Qi~She, and Shanghang Zhang. 2024{\natexlab{b}}.
\newblock [cls] attention is all you need for training-free visual token pruning: Make vlm inference faster.
\newblock \emph{arXiv preprint arXiv:2412.01818}.

\bibitem[{Zhang et~al.(2024{\natexlab{c}})Zhang, Wu, Li, Li, Ma, Liu, and Li}]{zhang2024video}
Yuanhan Zhang, Jinming Wu, Wei Li, Bo~Li, Zejun Ma, Ziwei Liu, and Chunyuan Li. 2024{\natexlab{c}}.
\newblock Video instruction tuning with synthetic data.
\newblock \emph{arXiv preprint arXiv:2410.02713}.

\bibitem[{Zhang et~al.(2023)Zhang, Sheng, Zhou, Chen, Zheng, Cai, Song, Tian, R{\'e}, Barrett et~al.}]{zhang2023h2o}
Zhenyu Zhang, Ying Sheng, Tianyi Zhou, Tianlong Chen, Lianmin Zheng, Ruisi Cai, Zhao Song, Yuandong Tian, Christopher R{\'e}, Clark Barrett, and 1 others. 2023.
\newblock H2o: Heavy-hitter oracle for efficient generative inference of large language models.
\newblock \emph{Advances in Neural Information Processing Systems}, 36:34661--34710.

\bibitem[{Zhong et~al.(2024)Zhong, Liu, Li, and Wang}]{zhong2024aim}
Yiwu Zhong, Zhuoming Liu, Yin Li, and Liwei Wang. 2024.
\newblock Aim: Adaptive inference of multi-modal llms via token merging and pruning.
\newblock \emph{arXiv preprint arXiv:2412.03248}.

\bibitem[{Zhou et~al.(2024)Zhou, Shu, Zhao, Wu, Xiao, Yang, Xiong, Zhang, Huang, and Liu}]{zhou2024mlvu}
Junjie Zhou, Yan Shu, Bo~Zhao, Boya Wu, Shitao Xiao, Xi~Yang, Yongping Xiong, Bo~Zhang, Tiejun Huang, and Zheng Liu. 2024.
\newblock Mlvu: A comprehensive benchmark for multi-task long video understanding.
\newblock \emph{arXiv preprint arXiv:2406.04264}.

\bibitem[{Zhu et~al.(2023)Zhu, Chen, Shen, Li, and Elhoseiny}]{zhu2023minigpt}
Deyao Zhu, Jun Chen, Xiaoqian Shen, Xiang Li, and Mohamed Elhoseiny. 2023.
\newblock Minigpt-4: Enhancing vision-language understanding with advanced large language models.
\newblock \emph{arXiv preprint arXiv:2304.10592}.

\end{thebibliography}

\appendix

\section{Appendix}
\label{sec:appendix}

\subsection{Baseline Settings}
\label{subsec:experimantal_settings}
For VisionZip, we set the ratio of dominant visual tokens and contextual visual tokens to 27:5 following the default setting. In LLaVA-Video~\cite{zhang2024video}, to solve the existence of newline tokens, after visual encoder, we first applied original method in VisionZip, then added newline tokens following, making sure the ratio of compressed visual tokens and added newline tokens remains fixed.

For Dycoke, L is set to 3, P is set to 0.7 following the default settings, and K is set to 0.7, 0.9 and 0.95 for fair comparison with three compression ratios $p=2.0,1.5,1.0$ adopted in our experiments.

For FastV, following the default settings, we couduct token pruning at layer 2 for all Video LLMs. The visual token retention ratio is set to $0.50, 0.375$ and $0.25$ for three compression ratios $p=2.0,1.5,1.0$ adopted in our experiments.

\subsection{Ablation studies on hyperparameters}
\label{subsec:ablation-hyperparameters}
% For progressive layer-by-layer token pruning, we start pruning from Layer 2 while keeping the last 12 layers unpruned across all Video LLMs. A uniform pruning rate is applied between the pruned layers, with a fixed number of tokens discarded after each layer. In LLaVA-OneVision-0.5B~\cite{Li2024LLaVAOneVision}, which consists of 24 layers, token pruning is by default performed from Layer 2 to Layer 12. Variants with different pruning depths are also presented for comparison.

\begin{table*}[t]
\centering
\captionsetup{font=small, skip=4pt}
\caption{Ablation study on different hyperparameters.}
\label{tab:ablation-all}

\begin{subtable}[t]{0.48\linewidth}
\centering
\caption{Ablation on $\tau_t$}
\label{tab:ablation-lambda1}
\resizebox{\linewidth}{!}{
\begin{tabular}{ccccc}
\toprule
\textbf{$\tau_t$} & VideoMME & MLVU & LongVideoBench & Avg \\
\midrule
0.03 & 44.63 & 46.40 & 46.30 & 45.78 \\
0.04 & 44.81 & 47.35 & 46.37 & 46.18 \\
\textbf{0.05} & \textbf{44.96} & \textbf{47.58} & \textbf{47.05} & \textbf{46.53} \\
0.06 & 44.85 & 47.50 & 46.37 & 46.24 \\
0.07 & 44.70 & 47.47 & 46.97 & 46.38 \\
\bottomrule
\end{tabular}
}
\end{subtable}
\hfill
\begin{subtable}[t]{0.48\linewidth}
\centering
\caption{Ablation on $\alpha_{\text{boost}}$}
\label{tab:ablation-lambda2}
\resizebox{\linewidth}{!}{
\begin{tabular}{ccccc}
\toprule
\textbf{$\alpha_{\text{boost}}$} & VideoMME & MLVU & LongVideoBench & Avg \\
\midrule
1.6 & 44.44 & 47.28 & 47.19 & 46.30 \\
1.8 & 44.44 & 47.51 & 47.05 & 46.33 \\
2.0 & 44.74 & \textbf{47.66} & 46.75 & 46.38 \\
\textbf{2.2} & \textbf{44.96} & 47.58 & 47.05 & \textbf{46.53} \\
2.4 & 44.56 & 47.49 & \textbf{47.27} & 46.44 \\
2.6 & 44.37 & 47.52 & 46.97 & 46.29 \\
\bottomrule
\end{tabular}
}
\end{subtable}

\vspace{0.3cm}

\begin{subtable}[t]{0.48\linewidth}
\centering
\caption{Ablation on $\gamma_{\text{cap}}$}
\label{tab:ablation-lambda3}
\resizebox{\linewidth}{!}{
\begin{tabular}{ccccc}
\toprule
\textbf{$\gamma_{\text{cap}}$} & VideoMME & MLVU & LongVideoBench & Avg \\
\midrule
0.50 & 44.81 & 46.80 & \textbf{47.05} & 46.22 \\
\textbf{0.75} & \textbf{44.96} & \textbf{47.58} & \textbf{47.05} & \textbf{46.53} \\
1.00 & 44.89 & 47.23 & 46.82 & 46.35 \\
\bottomrule
\end{tabular}
}
\end{subtable}
\hfill
\begin{subtable}[t]{0.48\linewidth}
\centering
\caption{Ablation on $\tau_s$}
\label{tab:ablation-tau_s}
\resizebox{\linewidth}{!}{
\begin{tabular}{ccccc}
\toprule
\textbf{$\tau_s$} & VideoMME & MLVU & LongVideoBench & Avg \\
\midrule
0.60 & 38.00 & 43.32 & 43.16 & 41.49 \\
0.80 & 41.93 & 44.81 & 45.85 & 44.20 \\
0.90 & 44.11 & 47.44 & 46.67 & 46.07 \\
\textbf{0.95} & \textbf{44.96} & \textbf{47.58} & 47.05 & \textbf{46.53} \\
1.00 & 44.63 & 47.24 & \textbf{47.19} & 46.31 \\
\bottomrule
\end{tabular}
}
\end{subtable}
\end{table*}

$\tau_t$ sets the threshold for significant segments. Values that are too high or too low can lead to segments more relevant to the text not being accurately distinguished; $\alpha_{\text{boost}}$ and $\gamma_{\text{cap}}$ control the number of tokens to keep within each significant segment. We properly adjusted their values to retain sufficient but not redundant tokens in significant segments, while keeping necessary tokens in others; $\tau_s$ influences the segment partitioning process. If too high, similar frames may be separated into different segments; if too low, segments may contain cluttered visual information and discard useful visual tokens. The results are shown in Tab.~\ref{tab:ablation-lambda1}--\ref{tab:ablation-tau_s}.

It is worth noting that our pipeline demonstrates strong performance across a wide range of hyperparameter settings, highlighting its robustness.

\subsection{Additional Visualizations of attention bias phenomenon}
\label{subsec:more-vis-layers}
\begin{figure*}[htbp]
    \centering

    % 第一行
    \begin{minipage}[b]{0.24\textwidth}
        \includegraphics[width=\linewidth]{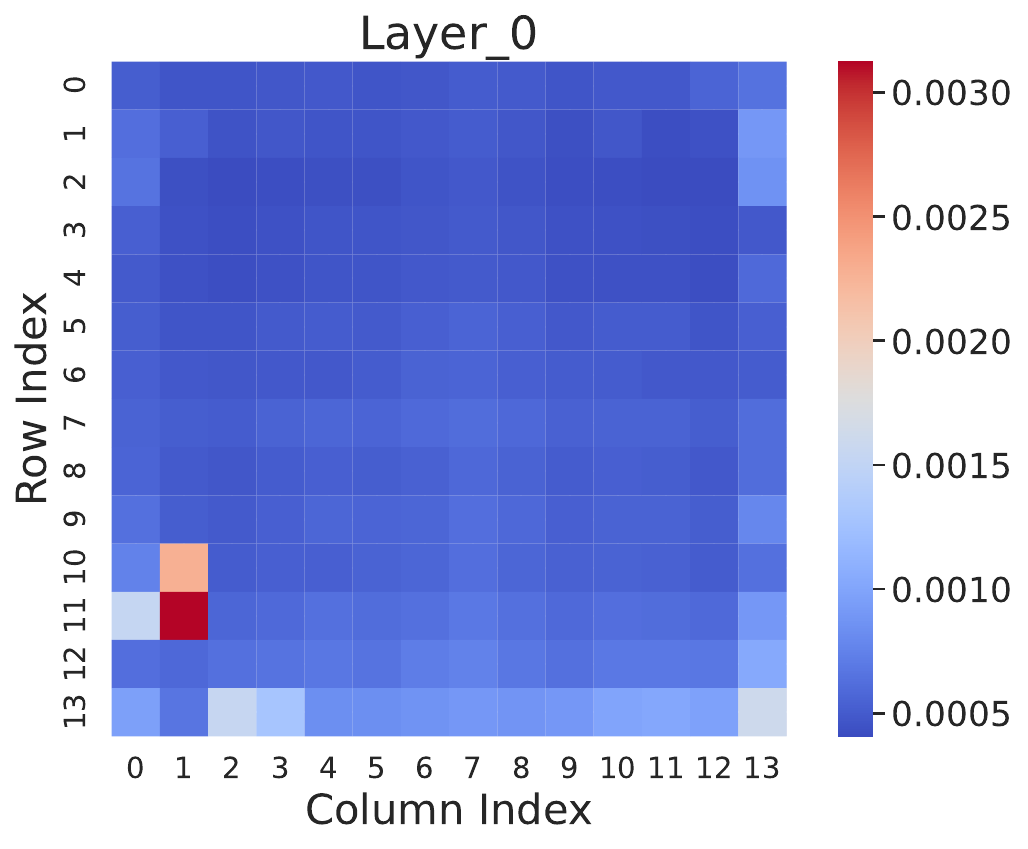}
    \end{minipage}
    \begin{minipage}[b]{0.24\textwidth}
        \includegraphics[width=\linewidth]{latex/figures/local_1.pdf}
    \end{minipage}
    \vspace{0.2cm}
    \begin{minipage}[b]{0.24\textwidth}
        \includegraphics[width=\linewidth]{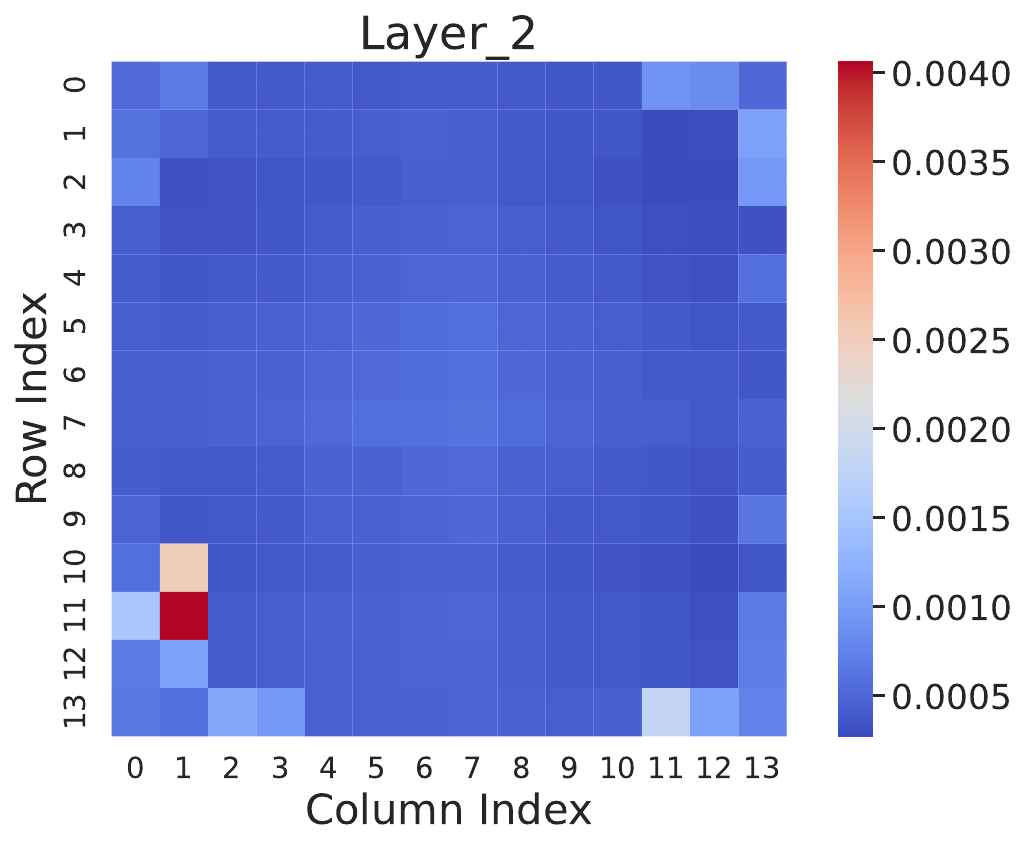}
    \end{minipage}
    \begin{minipage}[b]{0.24\textwidth}
        \includegraphics[width=\linewidth]{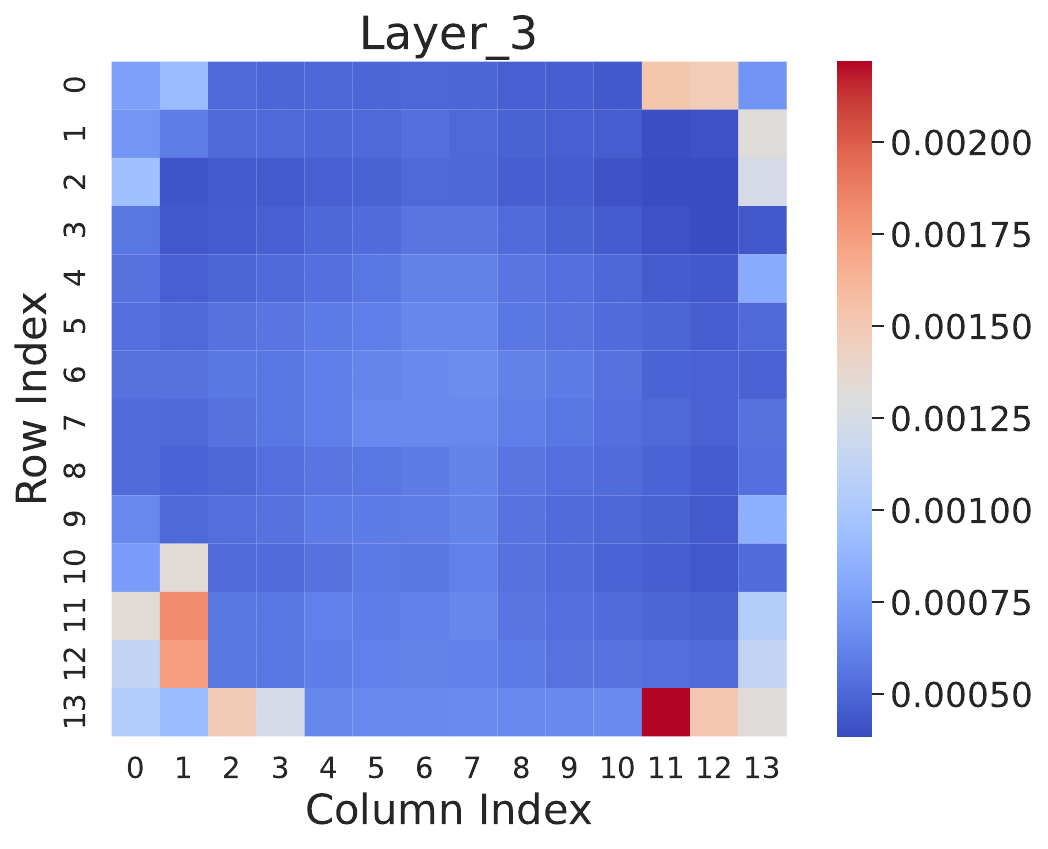}
    \end{minipage}

    \vspace{0.2cm}

    % 第二行
    \begin{minipage}[b]{0.24\textwidth}
        \includegraphics[width=\linewidth]{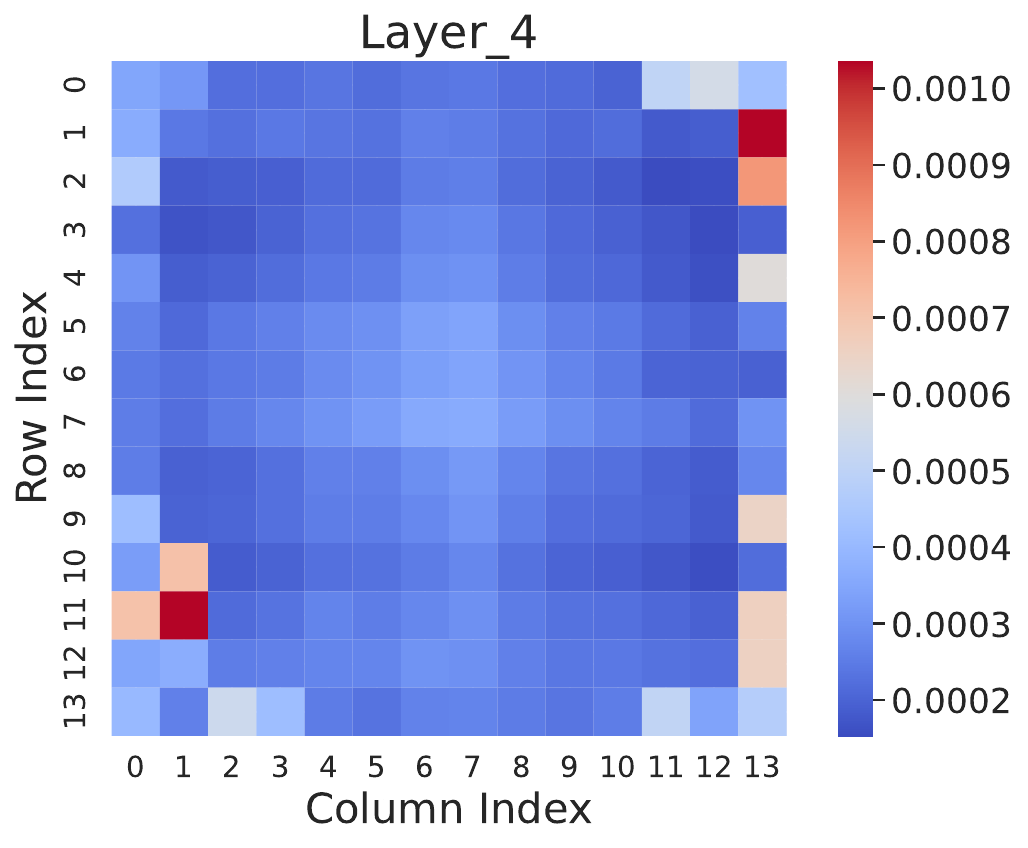}
    \end{minipage}
    \begin{minipage}[b]{0.24\textwidth}
        \includegraphics[width=\linewidth]{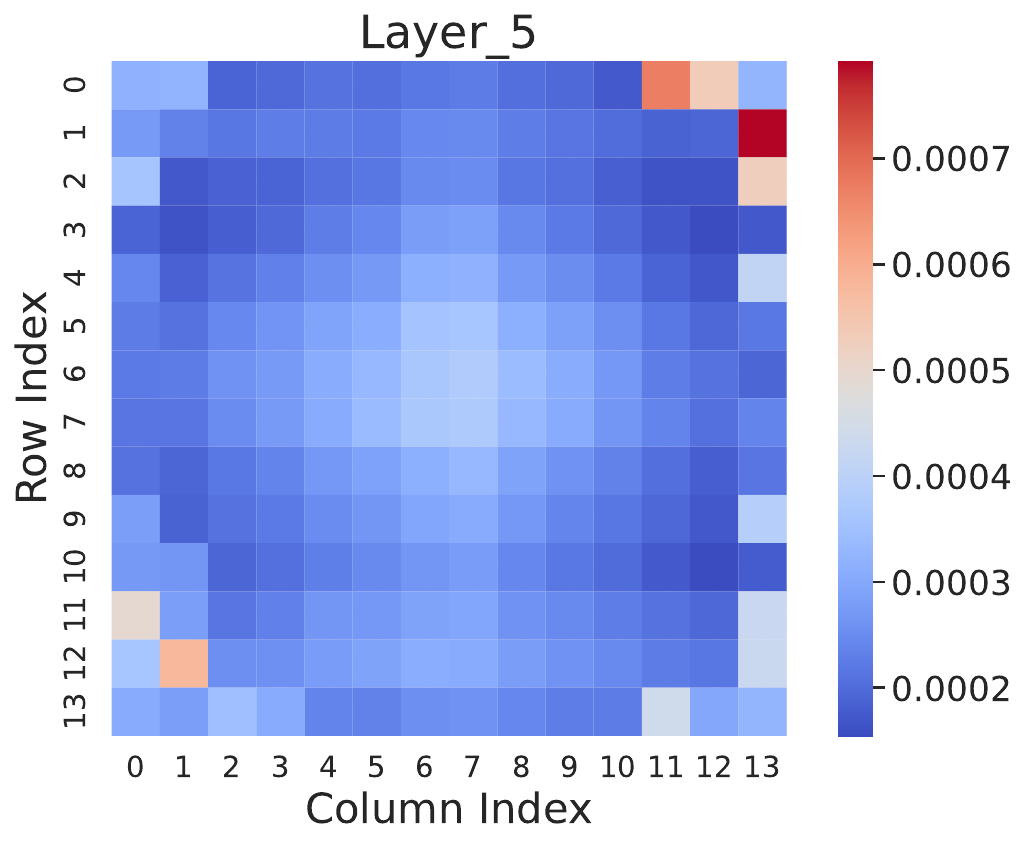}
    \end{minipage}
    \begin{minipage}[b]{0.24\textwidth}
        \includegraphics[width=\linewidth]{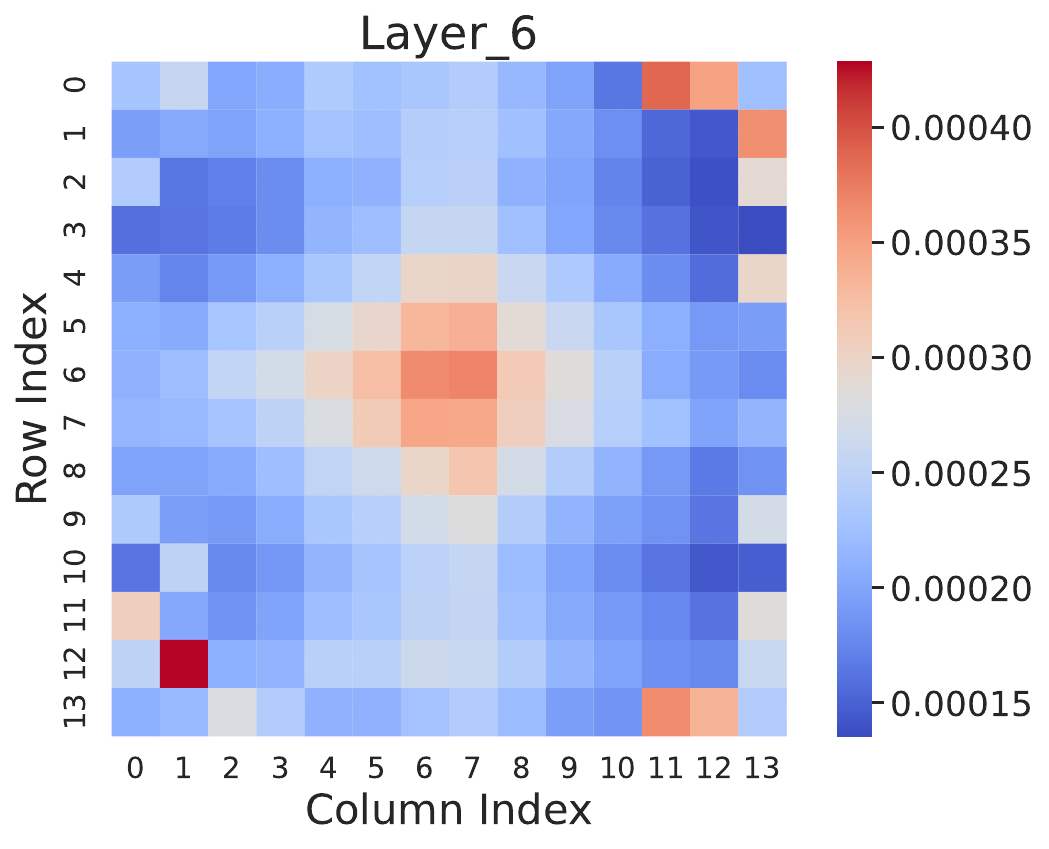}
    \end{minipage}
    \begin{minipage}[b]{0.24\textwidth}
        \includegraphics[width=\linewidth]{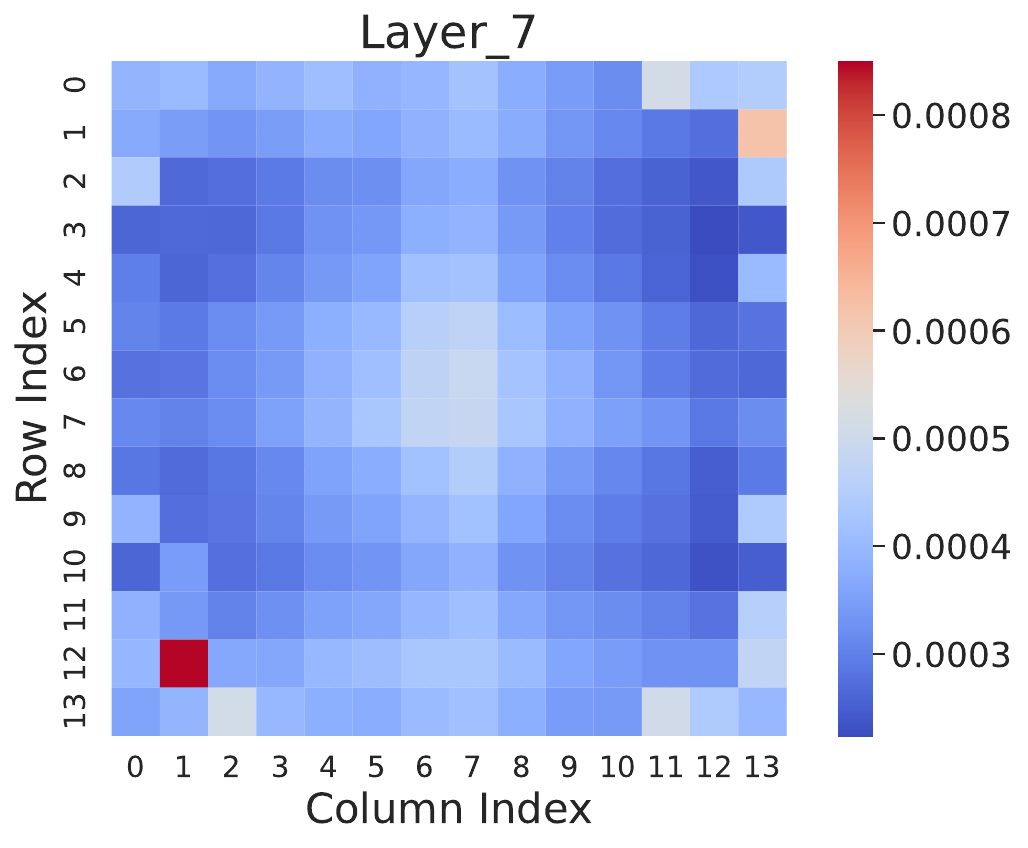}
    \end{minipage}

    \vspace{0.2cm}

    % 第二行
    \begin{minipage}[b]{0.24\textwidth}
        \includegraphics[width=\linewidth]{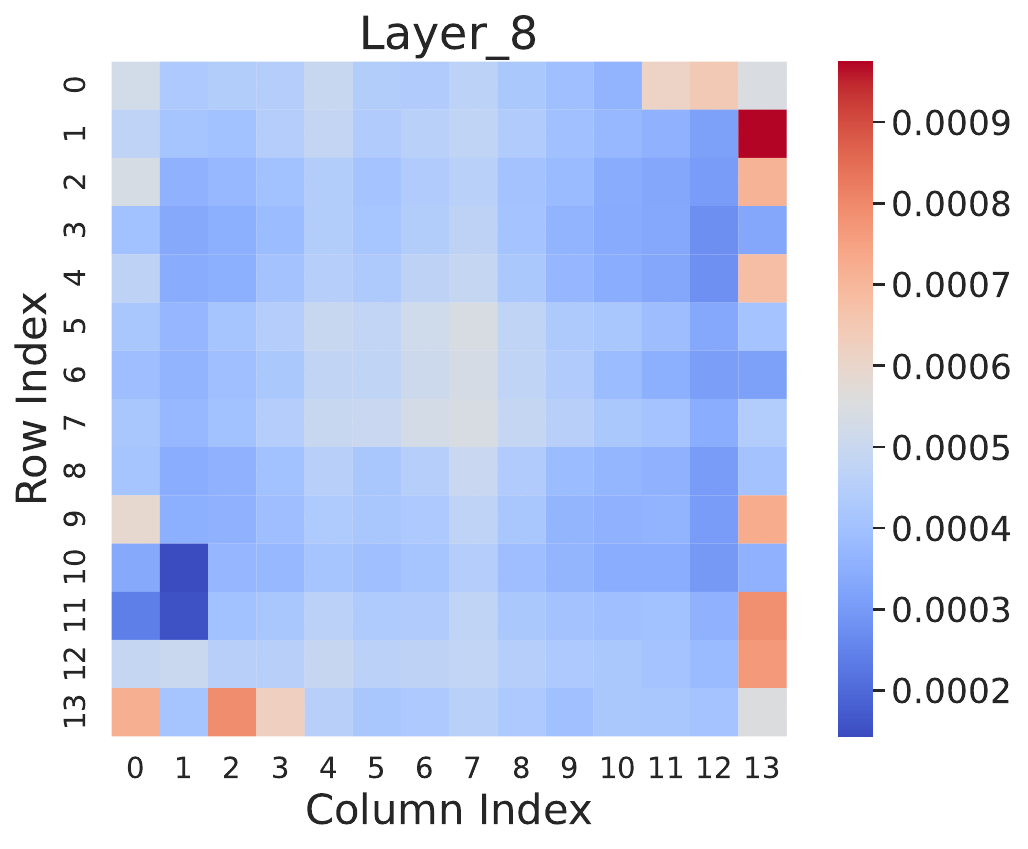}
    \end{minipage}
    \begin{minipage}[b]{0.24\textwidth}
        \includegraphics[width=\linewidth]{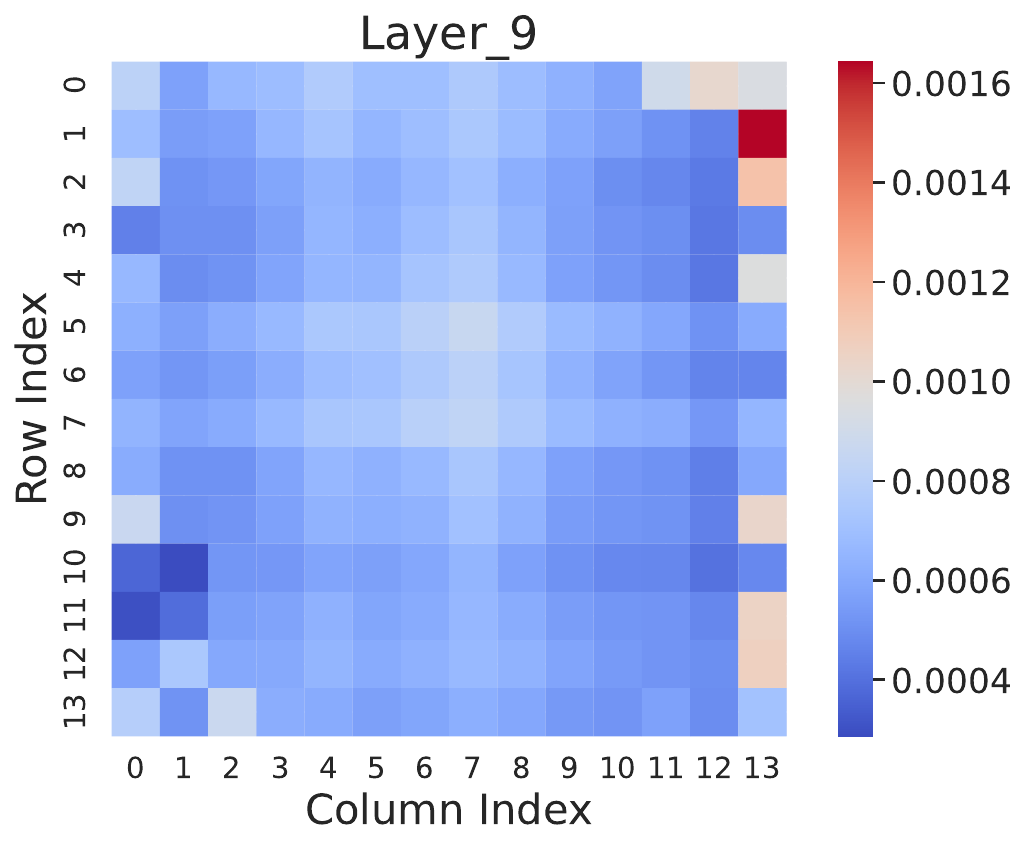}
    \end{minipage}
    \begin{minipage}[b]{0.24\textwidth}
        \includegraphics[width=\linewidth]{latex/figures/local_10.pdf}
    \end{minipage}
    \begin{minipage}[b]{0.24\textwidth}
        \includegraphics[width=\linewidth]{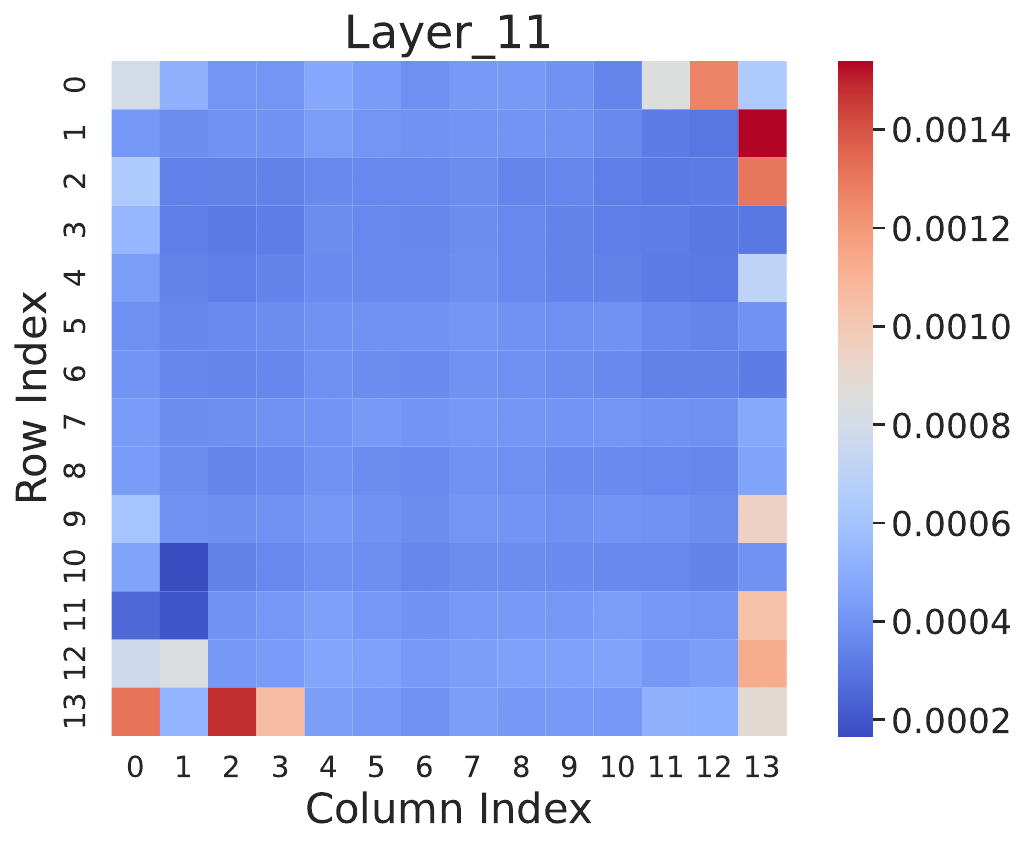}
    \end{minipage}

    \vspace{0.2cm}

    % 第二行
    \begin{minipage}[b]{0.24\textwidth}
        \includegraphics[width=\linewidth]{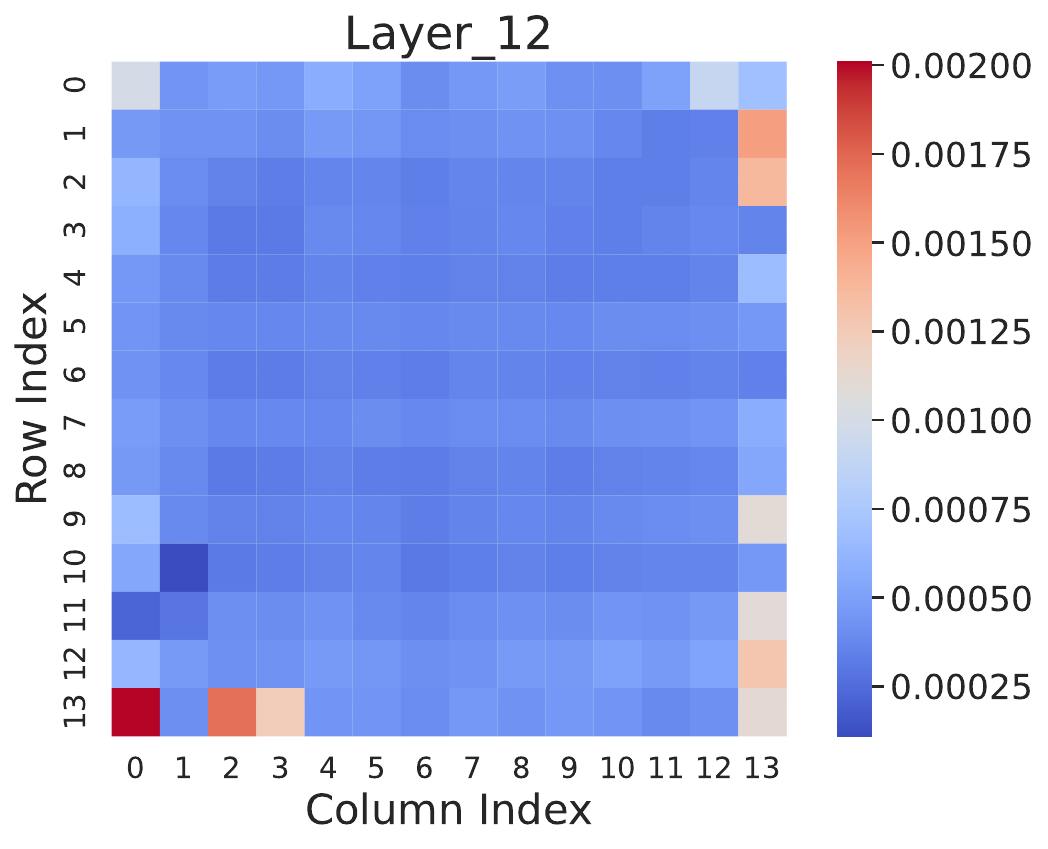}
    \end{minipage}
    \begin{minipage}[b]{0.24\textwidth}
        \includegraphics[width=\linewidth]{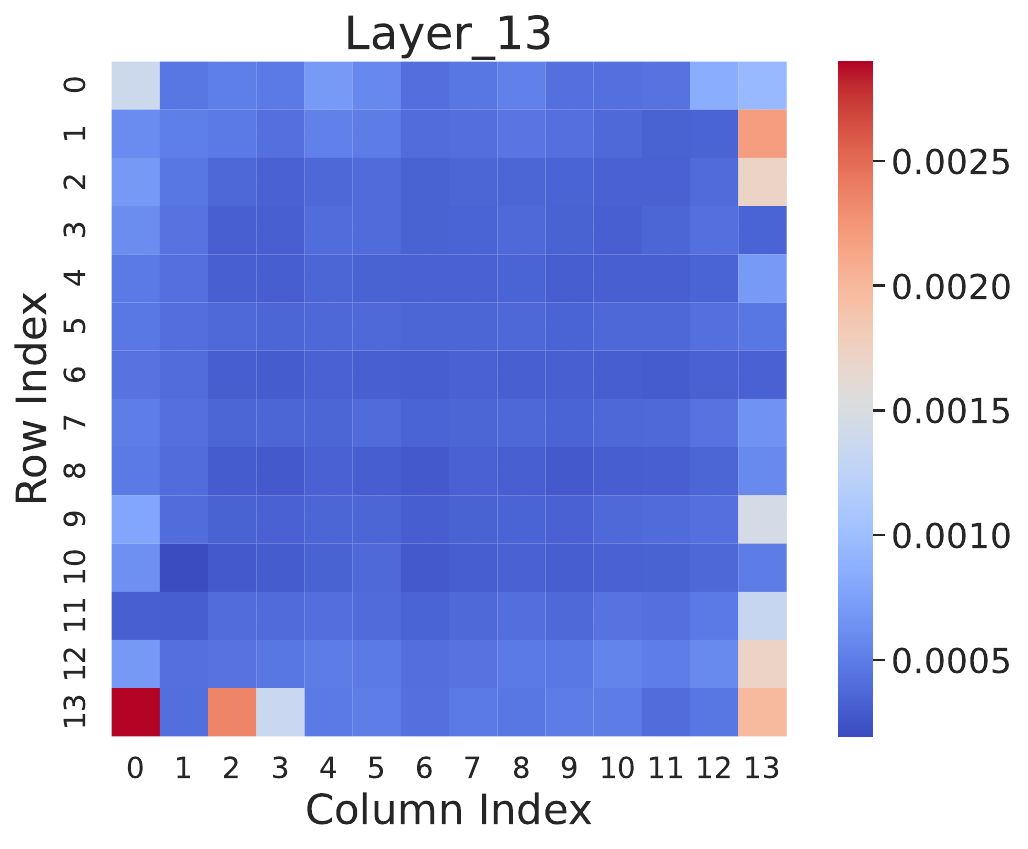}
    \end{minipage}
    \begin{minipage}[b]{0.24\textwidth}
        \includegraphics[width=\linewidth]{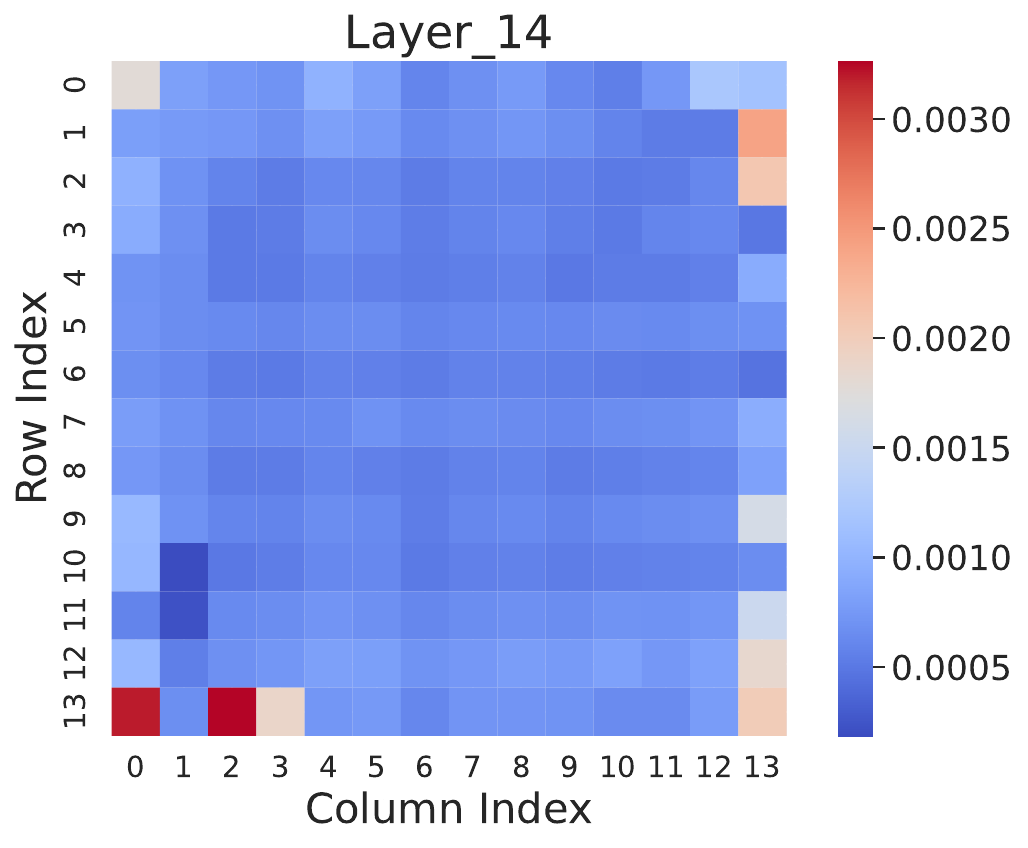}
    \end{minipage}
    \begin{minipage}[b]{0.24\textwidth}
        \includegraphics[width=\linewidth]{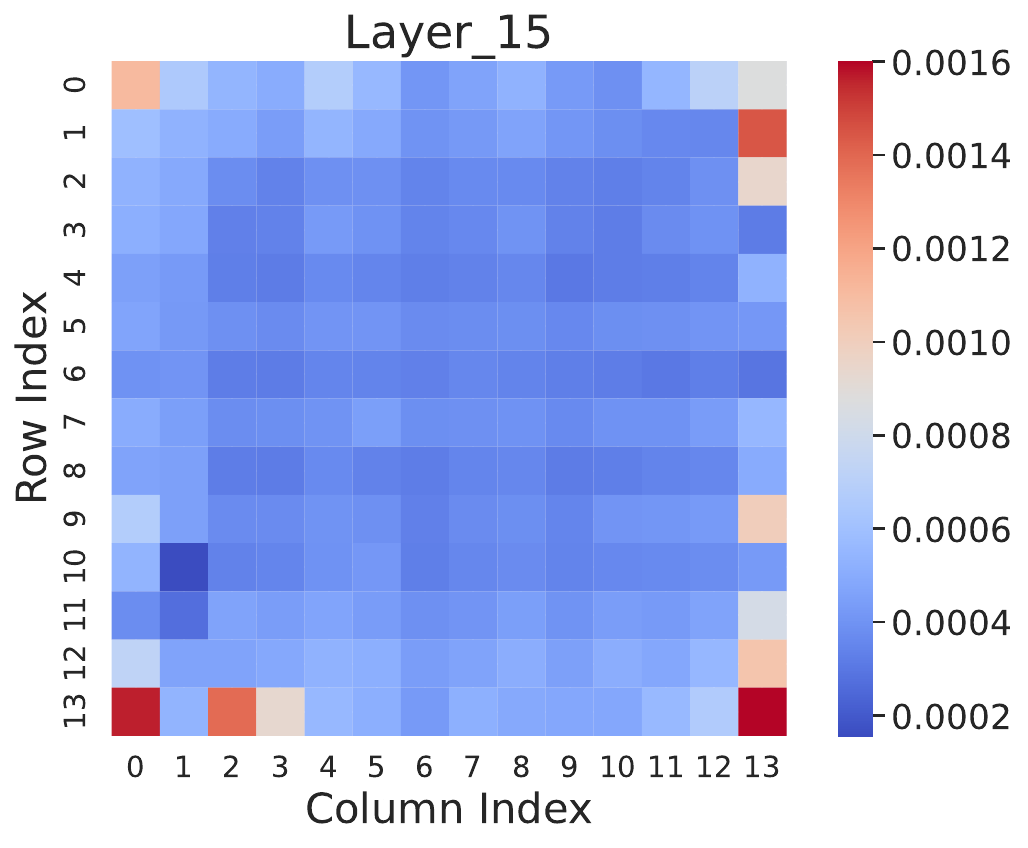}
    \end{minipage}

    \caption{Additional visualizations of local attention bias across different layers.}
    \label{fig:appendix_attention}
\end{figure*}

\begin{figure*}[htbp]
    \centering

    \begin{minipage}[b]{0.48\textwidth}
        \includegraphics[width=\linewidth]{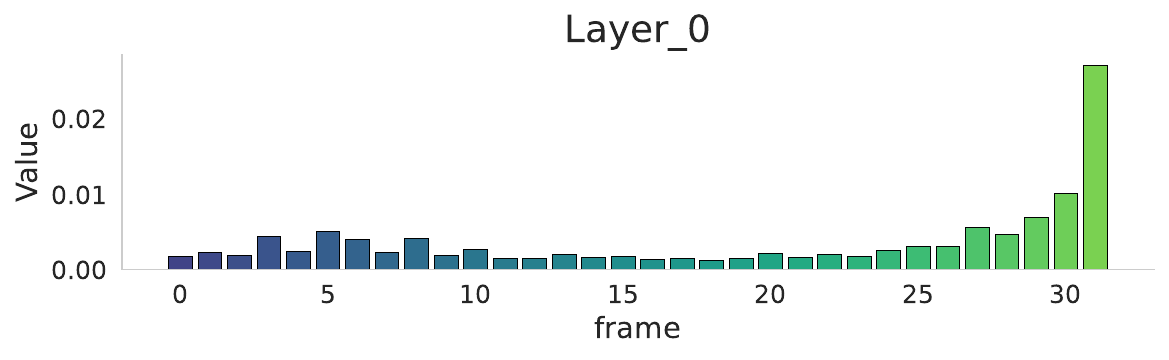}
    \end{minipage}
    \begin{minipage}[b]{0.48\textwidth}
        \includegraphics[width=\linewidth]{latex/figures/frame_1.pdf}
    \end{minipage}

    \vspace{0.2cm}

    \begin{minipage}[b]{0.48\textwidth}
        \includegraphics[width=\linewidth]{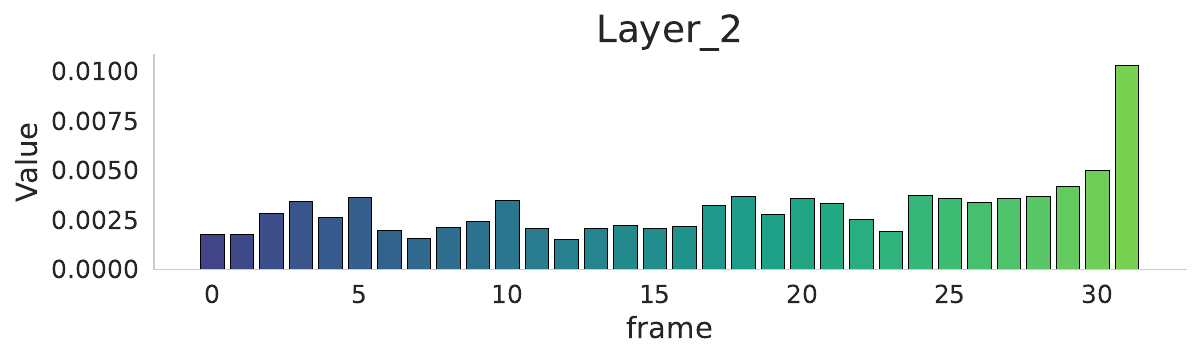}
    \end{minipage}
    \begin{minipage}[b]{0.48\textwidth}
        \includegraphics[width=\linewidth]{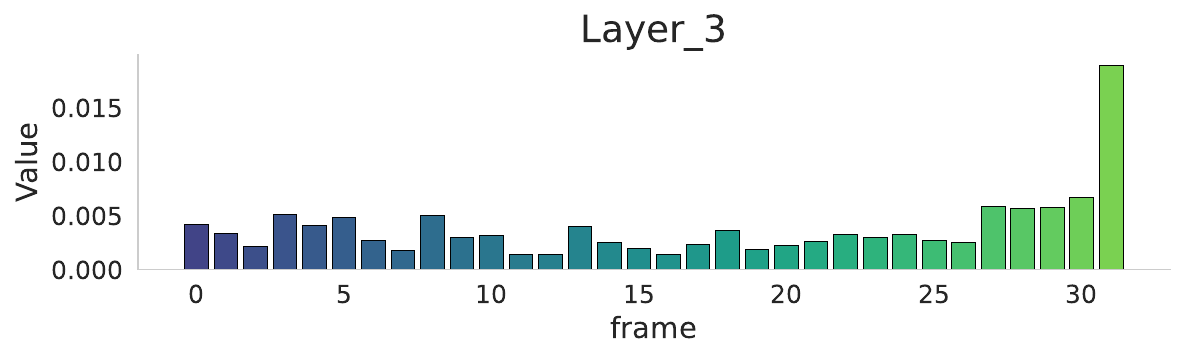}
    \end{minipage}

    \vspace{0.2cm}

    \begin{minipage}[b]{0.48\textwidth}
        \includegraphics[width=\linewidth]{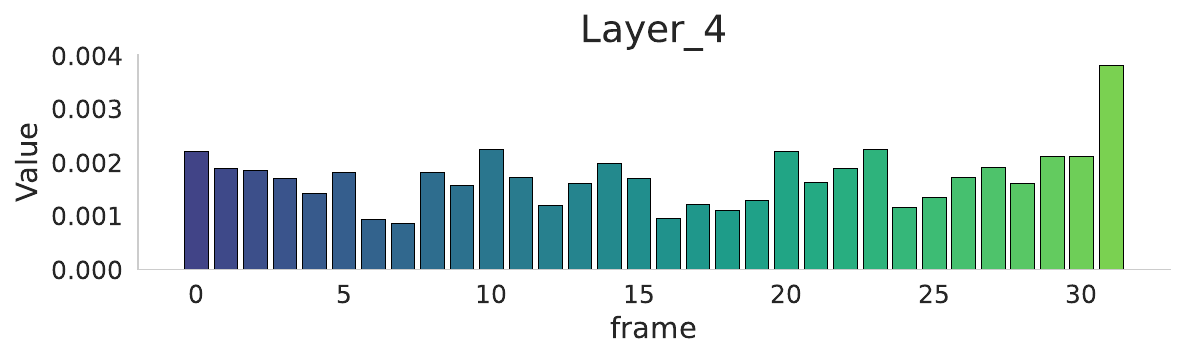}
    \end{minipage}
    \begin{minipage}[b]{0.48\textwidth}
        \includegraphics[width=\linewidth]{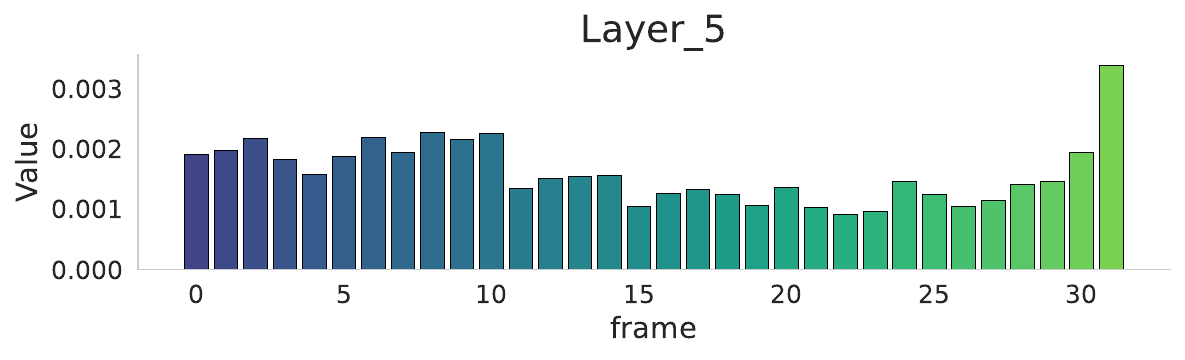}
    \end{minipage}

    \vspace{0.2cm}

    \begin{minipage}[b]{0.48\textwidth}
        \includegraphics[width=\linewidth]{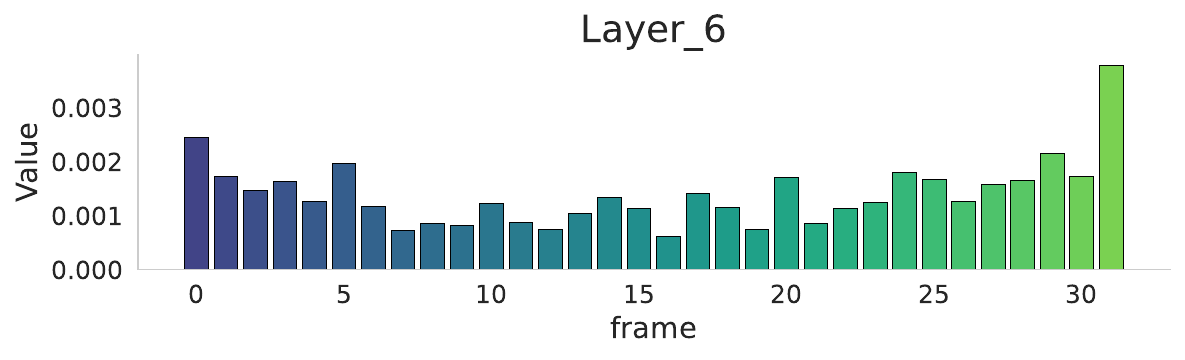}
    \end{minipage}
    \begin{minipage}[b]{0.48\textwidth}
        \includegraphics[width=\linewidth]{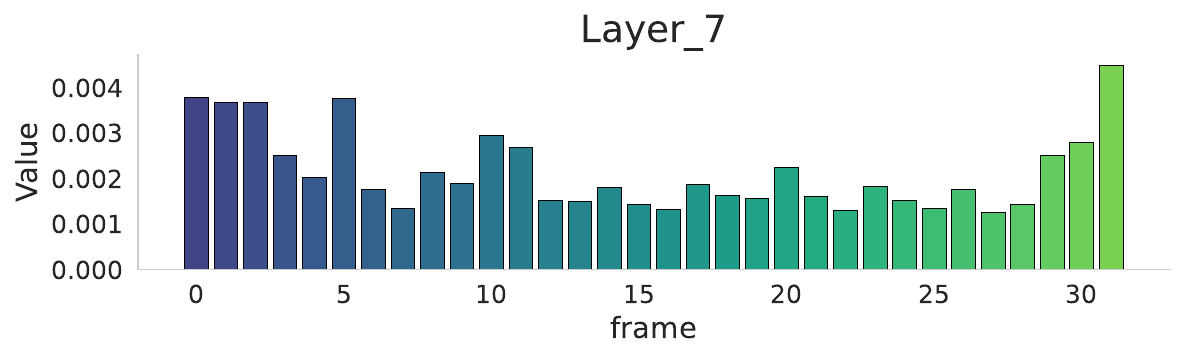}
    \end{minipage}

    \vspace{0.2cm}

    \begin{minipage}[b]{0.48\textwidth}
        \includegraphics[width=\linewidth]{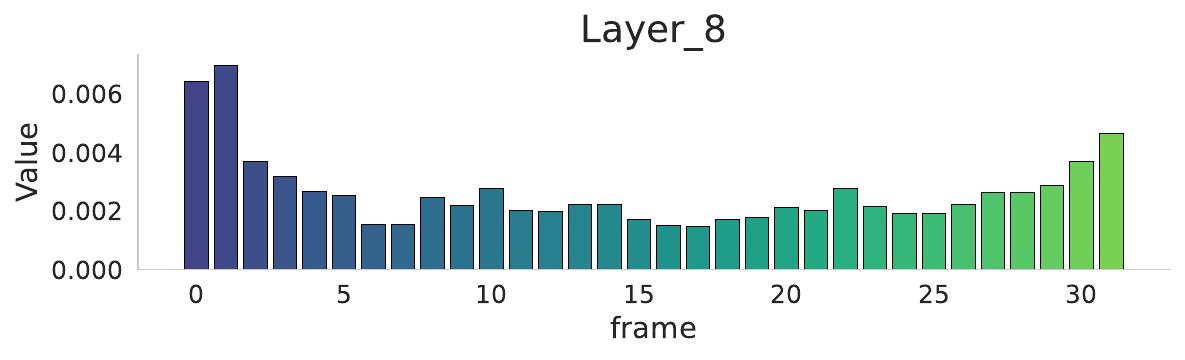}
    \end{minipage}
    \begin{minipage}[b]{0.48\textwidth}
        \includegraphics[width=\linewidth]{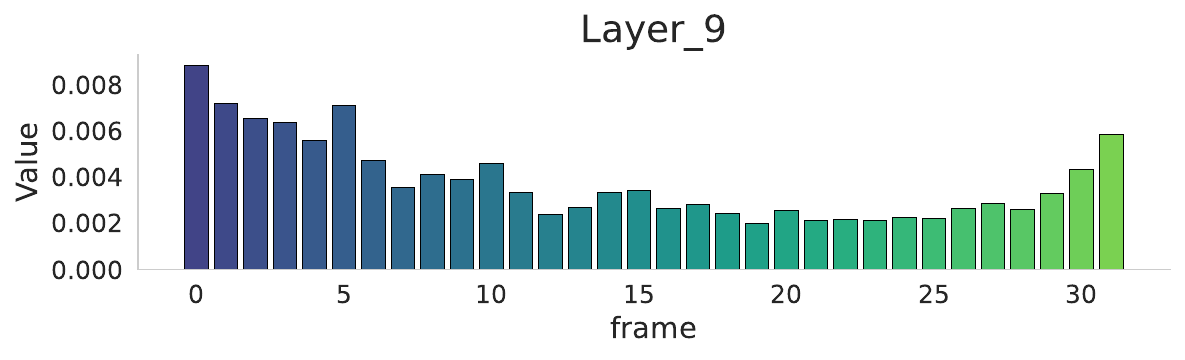}
    \end{minipage}

    \vspace{0.2cm}

    \begin{minipage}[b]{0.48\textwidth}
        \includegraphics[width=\linewidth]{latex/figures/frame_10.pdf}
    \end{minipage}
    \begin{minipage}[b]{0.48\textwidth}
        \includegraphics[width=\linewidth]{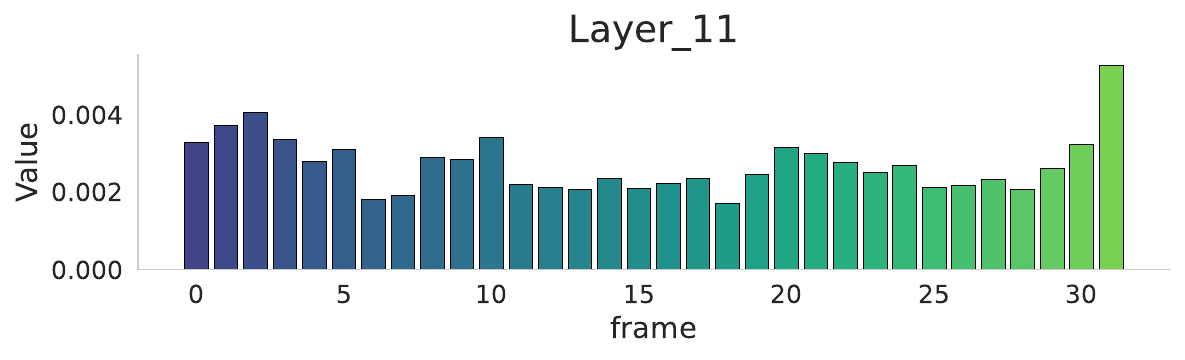}
    \end{minipage}

    \vspace{0.2cm}

    \begin{minipage}[b]{0.48\textwidth}
        \includegraphics[width=\linewidth]{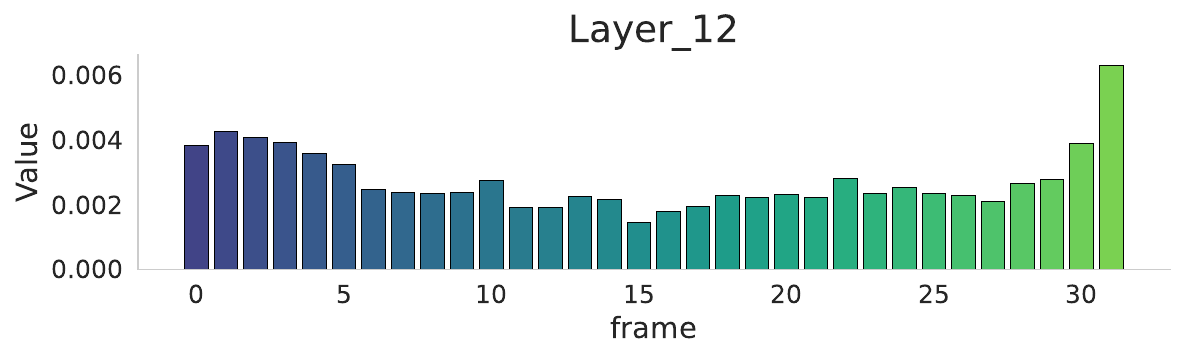}
    \end{minipage}
    \begin{minipage}[b]{0.48\textwidth}
        \includegraphics[width=\linewidth]{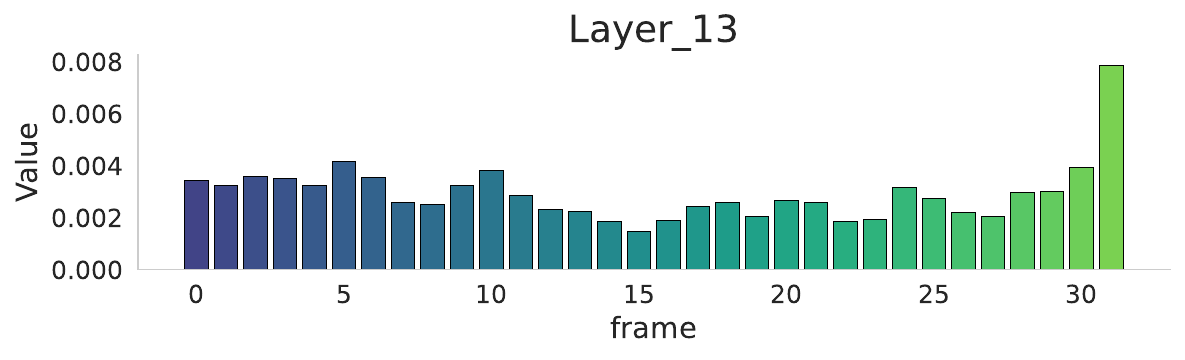}
    \end{minipage}

    \vspace{0.2cm}
    \begin{minipage}[b]{0.48\textwidth}
        \includegraphics[width=\linewidth]{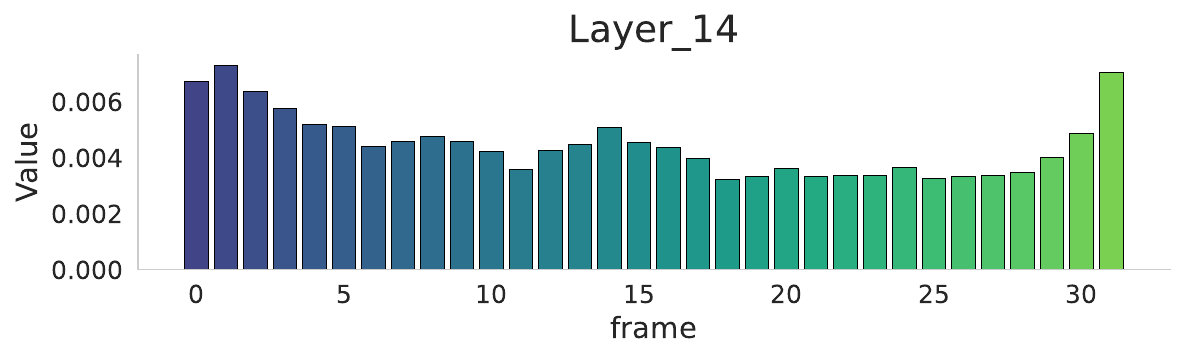}
    \end{minipage}
    \begin{minipage}[b]{0.48\textwidth}
        \includegraphics[width=\linewidth]{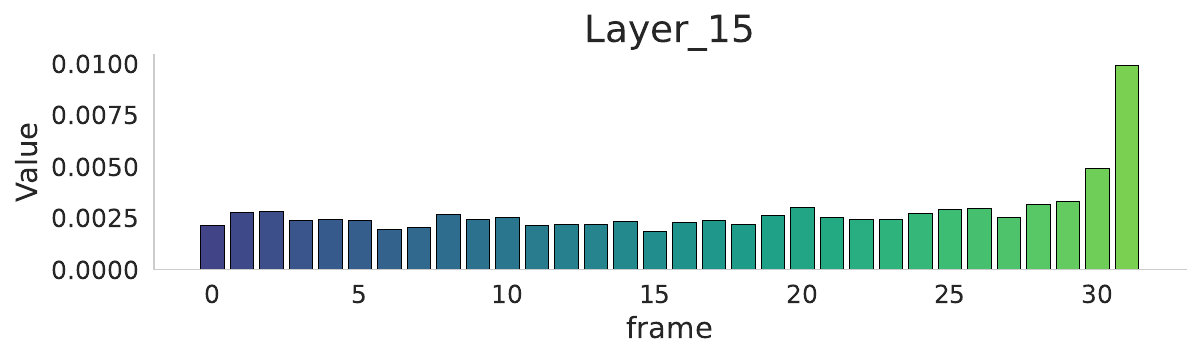}
    \end{minipage}
    \caption{Additional visualizations of global attention bias across different layers.}
    \label{fig:appendix_attention_2}
\end{figure*}

In Fig.~\ref{fig:combined_heatmaps_frames}--\ref{fig:position_heatmap_all}, we visualized the attention bias phenomenon in layer 1 and 10 from LLaVA-OneVision-7B~\cite{Li2024LLaVAOneVision}. More visualizations from other layers are shown in Fig.~\ref{fig:appendix_attention} and Fig.~\ref{fig:appendix_attention_2}. It can be observed that the global and local attention bias we mentioned in this paper appear across all the layers shown, demonstrating the universality and consistency of our observation.

\end{document}